\documentclass{article}

\usepackage{arxiv}

\usepackage[utf8]{inputenc} 
\usepackage[T1]{fontenc}    
\usepackage{hyperref}       
\usepackage{url}            
\usepackage{booktabs}       
\usepackage{amsfonts}       
\usepackage{nicefrac}       
\usepackage{microtype}      
\usepackage{cleveref}       
\usepackage{lipsum}         
\usepackage{graphicx}
\usepackage{natbib}
\usepackage{doi}
\usepackage{float}
\usepackage{CJKutf8}
\usepackage{tikz}
\usepackage{listings}
\usepackage{multirow}
\usepackage{array}
\usepackage[skip=5pt]{caption}
\usepackage{pgfplots} 
\pgfplotsset{compat=newest}
\usepackage{changepage}

\def\StripPrefix#1>{}
\def\jobis#1{FF\fi
    \def\predicate{#1}%
    \edef\predicate{\expandafter\StripPrefix\meaning\predicate}%
    \edef\job{\jobname}%
    \ifx\job\predicate
}
\if\jobis{output}%
    \usepackage[frozencache, outputdir=../]{minted}
\else
    \usepackage[frozencache,cachedir=minted-cache]{minted}
\fi

\usetikzlibrary{arrows,shapes,positioning,shadows,trees,calc,shapes.multipart}

\title{TabLib: A Dataset of 627M Tables with Context}

\usepackage{authblk}

\author[]{%
	\hspace{1mm}Gus Eggert%
}
\author[]{%
	\hspace{1mm}Kevin Huo%
}
\author[]{%
	\hspace{1mm}Mike Biven%
}
\author[]{%
	\hspace{1mm}Justin Waugh%
}
\affil[]{
	Approximate Labs\thanks{\texttt{\href{mailto:research@approximatelabs.com}{research@approximatelabs.com}}},
	Boulder, CO\@, USA
}

\renewcommand{\shorttitle}{\textit{TabLib}}

\hypersetup{
pdftitle={TabLib: A Dataset of 627M Tables With Context},
pdfsubject={cs.CL},
pdfauthor={Gus Eggert, Kevin Huo, Mike Biven, Justin Waugh},
pdfkeywords={Tabular Data, LLM, Large Data Model, Dataset, Common Crawl, GitHub},
}

\begin{document}

\maketitle

\begin{abstract}
It is well-established that large, diverse datasets play a pivotal role in the performance of modern AI systems for text and image modalities. However,  there are no datasets for tabular data of comparable size and diversity to those available for text and images. Thus we present "TabLib'', a compilation of 627 million tables totaling 69 TiB, along with 867B tokens of context. TabLib was extracted from numerous file formats, including CSV, HTML, SQLite, PDF, Excel, and others, sourced from GitHub and Common Crawl. The size and diversity of TabLib offer considerable promise in the table modality, reminiscent of the original promise of foundational datasets for text and images, such as The Pile and LAION.
\end{abstract}

\section{Introduction}
The importance of data in model training has continued to grow \citep{hoffmann_training_2022}. Training data volume is now considered to be roughly as important to model performance as model size \citep{zha_data-centric_2023}. This implies that large datasets are promising assets for improving the performance of AI models. 

For example, in 2021 OpenAI released both CLIP and DALL-E \citep{radford_learning_2021,ramesh_zero-shot_2021}, which were considered state-of-the-art for image tasks. A large part of their success was due to their training data scale of 400M image-text pairs, whereas previously the largest open dataset for image-text pairs was around 10M \citep{schuhmann_laion-400m_2021}. Even larger training datasets such as LAION-5B \citep{schuhmann_laion-5b_2022} have fueled subsequent image models like Stable Diffusion \citep{rombach_high-resolution_2022}.

Given the volume and significance of information captured in tabular data, research on applying AI models to tabular data is an area of active research \citep{badaro_transformers_2023} \citep{jin_survey_2022} \citep{dong_table_2022}. Despite this, there are not many large-scale, diverse, and accessible datasets for tabular data. We are aware of only one large scale crawl that exceeds 10M tables (WebTables \citep{lehmberg_large_2016}), and only a few additional datasets have more than one million tables (WikiTables \citep{arenas_tabel_2015}, GitTables \citep{hulsebos_gittables_2023}, VizNet \citep{hu_viznet_2019}). Furthermore, the largest of these datasets (WebTables) is composed solely of HTML tables, which differ meaningfully from other common table types such as database tables, suggesting that WebTables may be insufficient for training models for diverse tasks. We believe that a larger and more diverse dataset will accelerate the advancement of tabular AI systems.

Thus, we present ``TabLib'', whose notable characteristics include:

\begin{itemize}
    \item \textbf{Scale}: Over 627 million individual tables totaling 69 TiB
    \item \textbf{Table metadata}: 867B tokens of contextual information, such as filenames, URLs, text before and after the table in the source document, and OpenGraph metadata.
    \item \textbf{Diversity}: Across language, category, size, source (Common Crawl\footnote{\url{https://commoncrawl.org}} and GitHub\footnote{\url{https://github.com}}), and format (CSV, HTML, PDF, Excel, SQLite, etc.)
    \item \textbf{Provenance}: Table source and transformation data to enable attribution and validation
\end{itemize}

These characteristics suggest TabLib could be a useful research asset for many fields, which we discuss later in \ref{sec:impact}\nameref{sec:impact}. We hope that TabLib will help advance tabular data understanding and catalyze the development of AI models focused on this modality, which we refer to as \emph{large data models}.

\subsection{Related Work}\label{sec:related_work}
Numerous open datasets exist for the purpose of training machine learning models to understand and interpret tabular data. Some of the most significant of these datasets are detailed in Table 2 in \citep{badaro_transformers_2023}. While high quality, existing datasets such as Spider, WikiDB, and VizNet \citep{vogel_wikidbs_2023, yu_spider_2019, hu_viznet_2019} lack the size and/or diversity necessary to pre-train large data models with broad applicability.

Two data sets have noteworthy volume: WebTables \citep{cafarella_webtables_2008} and GitTables \citep{hulsebos_gittables_2023}. 

The latest WebTables corpus contains 233 million tables extracted from HTML pages from Common Crawl \footnote{\url{https://webdatacommons.org/webtables/\#results-2015}}. WebTables contains a large volume of tables, but has limited diversity due to only including HTML tables from web pages.

GitTables is a continuously updated library of tables extracted from ``comma-separated value'' files (CSVs) hosted on GitHub, containing 1 million tables. These tables tend to be structurally different from the HTML-centric WebTables \citep{hulsebos_gittables_2023}, thus an important table corpus. Compared to WebTables, GitTables is relatively small, and still only supports a single file type (CSV).

\subsection{Impact}\label{sec:impact}

Applying AI to tabular data is an active field of study, and there are many applications and research areas that could significantly benefit from a large, diverse dataset such as TabLib. These include:

\begin{itemize}
	\item \textbf{Dataset Search:} Identifying corresponding tables using a set of keywords that describe the required information \citep{benjelloun_google_2020,chapman_dataset_2020,zhang_ad_2018}
	\item \textbf{Semantic Understanding:} Using data tables to create or augment general-purpose knowledge bases, and vice versa. \citep{dong_knowledge_2014,liu_tabular_2023,jimenez-ruiz_semtab_2020,efthymiou_matching_2017,bonfitto_table_2021, hulsebos_sherlock_2019} 
	\item \textbf{Data Integration:} Identifying tables that can be joined or unioned within a large corpus of tables. Includes schema mapping. \citep{dong_efficient_2021,zhang_recommending_2019,zhu_josie_2019,nargesian_table_2018,santos_correlation_2021,srinivas_lakebench_2023,zhu_auto-join_2017, cong_warpgate_2023, cong_pylon_2023}
	\item \textbf{Knowledge Extraction:} Interacting with data through natural language, via tasks like question answering and semantic parsing. \citep{zha_tablegpt_2023,cheng_is_2023,zhang_data-copilot_2023,li_can_2023,pourreza_din-sql_2023,talmor_multimodalqa_2021,lin_bridging_2020}
	\item \textbf{Table Metadata Prediction:} Predicting metadata such as column types, inclusion of personally identifiable information (PII), and data cleanliness. \citep{zhang_effective_2017,parikh_totto_2020,korini_column_2023}
 \item \textbf{Table Representation Learning:} Representing tables as a distinct modality of information for training machine learning models \citep{yin_tabert_2020, deng_turl_2020, tang_rpt_2021, herzig_tapas_2020, iida_tabbie_2021}
\end{itemize}

\section{Methods}\label{sec:collection}
\subsection{System Architecture}

We built a processing pipeline that consumes raw data from data sources, extracts tables into Pandas dataframes \citep{mckinney_data_2010}, serializes those dataframes into Arrow tables \footnote{\url{https://arrow.apache.org/}}, stores each in blob storage and metadata in a SQL database, and then aggregates into Parquet files \footnote{\url{https://parquet.apache.org/}}. To orchestrate this process, we used the Ray distributed processing framework \citep{moritz_ray_2018}.

Because parsing tabular data is relatively complex compared to text due to its additional structure and data types (see \nameref{sec:parsing}), we encountered some failure scenarios which were difficult to recover gracefully from, such as out-of-memory errors and catastrophic regular expression backtracking. As such, we isolated each ``source'' as its own task instead of batching them together. 

This granular task scheduling necessitated scheduling hundreds of millions of tasks. We found Ray's scheduler problematic for this, so we scheduled these tasks using a PostgreSQL database, and used Ray to maintain long-running tasks which pulled work from the database, extracted the tables and metadata, stored the tables in blob storage, and wrote the metadata back to the DB. A separate Ray actor tracked the progress of these tasks, handled timeouts and retries, occasionally aggregated batches of metadata into Parquet files, and wrote those into blob storage.

\begin{figure}[!ht]
	\centering
	\includegraphics[width=0.9\textwidth]{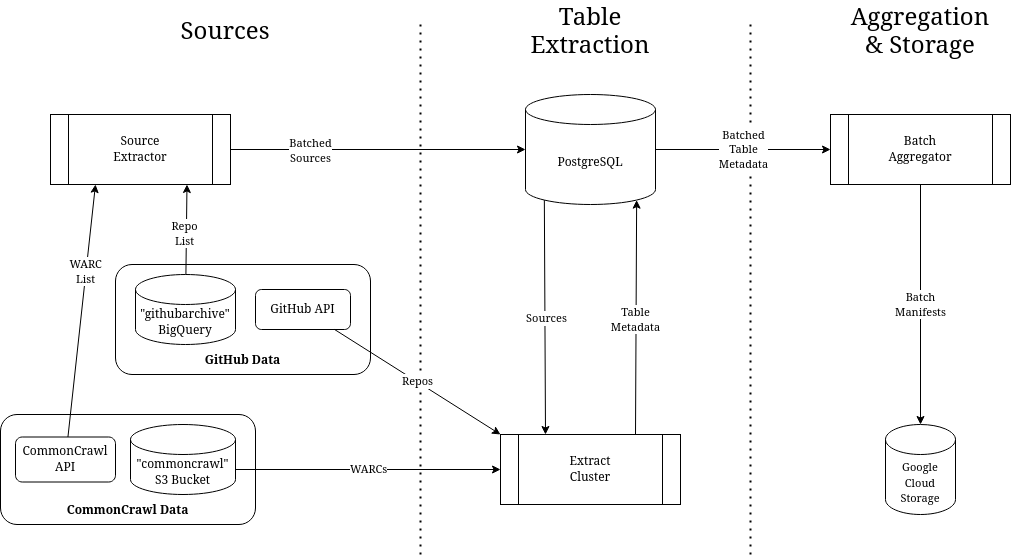}
	\caption{Architecture of table extraction pipeline.}
    \label{fig:extraction_pipeline}
\end{figure}

\subsection{Sources}\label{sec:sources}
For data about the number of tables extracted for each data source and file types, see \nameref{sec:summary-stats}. For samples of extracted tables and metadata, see \nameref{sec:sample_data} in the appendix.

\subsubsection{GitHub}

To reduce the amount of noise, we skipped all files under \verb|node_modules| directories, and all JSON and YAML files which are generally configuration files in GitHub. Since files in GitHub often contain extensions like \verb|.csv| that provide hints for the content type, we used Python's \verb|mimetypes.guess_type()| function to see if the file was a supported type; if not then we inspected the file's bytes using \verb|libmagic| \footnote{\url{https://www.darwinsys.com/file/}}, and if it was still unsupported then the file was skipped. Files larger than 1 GB were also skipped.

Tables extracted from GitHub repos result in the following fields in each table's \textbf{context\_metadata}:

\begin{itemize}
	\item \textbf{github\_repo}: the repo name
	\item \textbf{github\_ref}: the ref used, such as ``refs/heads/master''
	\item \textbf{github\_hash}: the shortened Git commit hash
	\item \textbf{github\_repo\_path}: the path of the file in the repo where the table was found
\end{itemize}

\subsubsection{Common Crawl}

We used the latest crawl at the time, which was \verb|CC-MAIN-2023-23|. Common Crawl results are serialized using the WARC format, which includes ``request'' and ``response'' records. We only considered response records. We discarded ``truncated'' responses which had response lengths that exceed Common Crawl's limit. If a WARC-Identified-Payload-Type record header was included in the record, then we used its mimetype as a hint for detecting the content type, otherwise we used the Content-Type header in the HTTP response, and followed a similar approach as GitHub (use the mimetype if possible, otherwise use libmagic). About 20\% of WARC files were dropped due to issues parsing certain HTML elements with Pandas.

Tables extracted from Common Crawl WARC records result in the following fields in each table's \textbf{context\_metadata}:

\begin{itemize}
    \item \textbf{warc\_path}: the path of the WARC file in Common Crawl
    \item \textbf{warc\_record\_id}: the record ID in the WARC file as specified by WARC-Record-ID
    \item \textbf{warc\_target\_uri}: the target URI of the HTTP request as specified by WARC-Target-URI
    \item \textbf{warc\_date}: the date of the request as specified by WARC-Date
\end{itemize}

\subsection{Storage Data Model}
In order to efficiently store and manage the large volume of tabular data in TabLib, we implemented a data storage model that consists of two main components: blob storage and manifests. Using this storage model, we can efficiently manage and retrieve the tables based on their metadata and content hash. This allows for easy deduplication, querying, and analysis of the dataset. A final post-processing step was performed which added the serialized tables as a column in the manifests, which is ultimately the TabLib schema, but this paper will focus on the intermediate representation because it is what the analyses are based on.

\subsubsection{Manifest Schema}\label{sec:manifest_schema}
The manifests contain metadata about the tables and are stored as partitioned Parquet files. The schema for the manifests includes the following fields:
\begin{itemize}
	\item \textbf{bucket}: the blob storage bucket of the table
	\item \textbf{key}: the blob storage key of the table
	\item \textbf{ref}: a human-readable string describing how the table was extracted
	\item \textbf{ref\_id}: a base64-encoded sha256 hash of the ref
	\item \textbf{exec\_id}: a UUIDv7 generated at the time of table extraction
	\item \textbf{run\_metadata}: serialized JSON object containing metadata about the run, including start and end times
	\item \textbf{context\_metadata}: serialized JSON object containing metadata about the table, including:
	      \begin{itemize}
		      \item \textbf{extractor}: the extractor used for this table (e.g. ``html'', ``csv'', ``pdf'', etc.)
		      \item \textbf{mime\_type}: the detected mime type of the bytes that the table was extracted from, e.g. ``text/html'' for an HTML page, ``text/csv'' for a CSV file, etc.
		      \item \textbf{<source-specific>}: additional fields depending on the source, see \nameref{sec:sources}
		      \item \textbf{<datatype-specific>}: additional fields depending on the data type, see \nameref{sec:parsing}
	      \end{itemize}
\end{itemize}

\subsubsection{Blob Storage Key Schema}\label{sec:blob_schema}

Each table in its intermediate form, before the final post-processing step, is stored as a separate blob object.  The blob's content is computed by serializing the Arrow table to bytes, and compressing these bytes with gzip. Each table is assigned a unique key based on the arrow table bytes content hash. The blob storage follows the following key schema:

\begin{itemize}
	\item \begin{verbatim}/manifests/{batch}/manifest.parquet\end{verbatim}
	\item \begin{verbatim}/tables/{batch}/{base64_sha256_of_arrow_table}\end{verbatim}
\end{itemize}

\subsection{Formats, Parsing, and Metadata}\label{sec:parsing}

Parsing tabular data presents unique challenges that are not present when parsing text. Tasks such as inferring column data types and row delimiters are complex and error-prone. Because of this, we reused existing open-source parsers as much as possible, such as those in Pandas and pdfplumber. For most file types, we drop parsed tables with only one column, one row, all empty column names, or only numeric column names.

Below we detail each data type and a summary of the parsing logic:

\begin{table}
	\centering
    \setlength\extrarowheight{2pt}
	\begin{tabular}{l m{9cm} l}
		\toprule
		\textbf{Data Type}                              & \textbf{Method} & \textbf{Context Metadata} \\
		\midrule

		HTML                                            &
		Parse with \verb|BeautifulSoup| using \verb|lxml| and \verb|html5lib| parsers.
		Then extract all \verb|<table>| elements with \verb|pandas.read_html()|.
		Extract HTML metadata with \verb|metadata_parser| library.
		Extract ``before'' and ``after'' context with \verb|BeautifulSoup|.
		Drop \verb|<table>| elements with \verb|colspan| values > 1000 to avoid causing out-of-memory errors.
		                                                &
		\begin{tabular}{@{}l@{}}
			\textbullet{} html\_title    \\
			\textbullet{} html\_metadata \\
			\textbullet{} before         \\
			\textbullet{} after          \\
		\end{tabular}                                                                  \\	\hline

		PDF                                             &
		Use \verb|pdfplumber| to extract tables.
		Only supports text-based tables and not image-based tables,
		and multi-page tables appear as a separate table per page.
		                                                &
		\begin{tabular}{@{}l@{}}
			\textbullet{} pdf\_bbox     \\
			\textbullet{} pdf\_page     \\
			\textbullet{} pdf\_metadata \\
			\textbullet{} before        \\
			\textbullet{} after         \\
		\end{tabular}                                                                   \\  \hline

		SQLite                                          &
		Use a custom SQLite VFS implementation to load the in-memory bytes with \verb|apsw|,
		list the tables, and then parse each table with \verb|pandas.read_sql()|.
		                                                &
		\begin{tabular}{@{}l@{}}
			\textbullet{} sqlite\_table         \\
			\textbullet{} sqlite\_other\_tables \\
		\end{tabular}                                                           \\ \hline

		Excel                                           &
		There are many Excel formats, mimetypes, and extensions,
		so ignore specifics and always try parsing as XLSX using \verb|openpyxl|,
		and then fall back to XLS  using \verb|xlrd|, using \verb|pandas.read_excel()|.
		Parse each sheet as its own table.
		                                                &
		n/a                                                                                           \\ \hline

		Parquet                                         &
		\verb|pandas.read_parquet()|                    &
		n/a                                                                                           \\ \hline

		JSON                                            &
		\verb|pandas.read_json(orient="records")|       &
		n/a                                                                                           \\ \hline

		YAML                                            &
		Use \verb|yaml.safe_load()| to convert to JSON,
		then \verb|pandas.read_json(orient="records")|. &
		n/a                                                                                           \\ \hline

		CSV                                             &
		\verb|pandas.read_csv(engine="python")|         &
		n/a                                                                                           \\ \hline

		TSV                                             &
		\verb|pandas.read_csv(engine="python")|         &
		n/a                                                                                           \\ \hline

		\bottomrule
	\end{tabular}
	\caption{\textbf{Summary of supported data types}, and how each was parsed.}
\end{table}

\section{Analysis and Results}

\subsection{Keys and Metadata}
\begin{table}[b]
    \centering
    \begin{tabular}{ll}
        \toprule
        Key Type 	& Number of Tables \\
        \midrule
        exec\_id 	& 660840556 \\
        \textbf{ref\_id} 		& 627208299 \\
        key 	& 459022959 \\
        content\_hash 		& 201376283 \\
        \bottomrule
    \end{tabular}
    \caption{\textbf{Unique counts of key-like values}, ordered by decreasing uniqueness. Each may be considered some definition of ``table''. We will use \texttt{ref\_id} as the definition of ``table'' for our analyses.}
    \label{tab:keys_breakdown}
\end{table}

We begin by examining the cardinalities of different keys: \verb|exec_id|, \verb|ref_id|, \verb|key|, and \verb|content_hash|, as shown in \autoref{tab:keys_breakdown}. Definitions of these values are in \nameref{sec:manifest_schema} and \nameref{sec:blob_schema}.

The \verb|exec_id| is unique across the dataset, generated upon line-item creation in the manifest. Any duplication indicates a serialization error. 

The \verb|ref_id| represents a unique source for a table. This should be unique across TabLib, but the current version of TabLib has some repetitions due to a bug in deduping items in the work queue. Future versions will allow tracking external data changes over time via \verb|ref_id|.

The number of unique \verb|key| values is substantial but not as large as unique \verb|ref_id| values. This discrepancy arises because the same content table can appear multiple times within a batch (e.g., a CSV file stored multiple times in a GitHub repository with different filenames). However, \verb|key| is not a global content-collision key as it includes the batch.

Table \verb|content_hash|es are 30.5\% the size of \verb|exec_id| values, indicating that most tables are not globally unique by content. The breakdown of these repeated tables is discussed further in \nameref{sec:duplication}.

For clarity, the term \emph{table} henceforth refers to a specific \verb|ref_id| instance. 

\subsection{Summary Statistics}\label{sec:summary-stats}

We calculate the total number of tables, total uncompressed table bytes, and total columns, broken out by data source and file type. See \autoref{tab:stats} for a summary of the dataset statistics.

\begin{table}[!ht]
    \scriptsize
	\centering
	\begin{tabular}{llllllll}
		\toprule
		Source 	& File Type	& Tables	& Bytes		& Columns 	& \multicolumn{3}{c}{Metadata Tokens} \\
		\cmidrule(lr){6-8} 
				& 			& 			& 			& 			& Ref 		& Column Names 	& Context Metadata \\
		\midrule
		\multirow{9}{*}{Common Crawl}& CSV 		& 90,667	& 3.10 GiB	& 2,265,499	& 15,630,941	& 12,488,110		& 17,984,442 \\
		& Excel 	& 143,012 	& 3.26 GiB 	& 1,836,491 & 21,063,876 & 10,579,484 	& 60,755,137 \\
		& HTML 	& 219,397,657 & 702.95 GiB 	& 1,076,171,440 & 29,602,279,431 & 3,686,722,801 	& 493,085,023,697 \\
		& JSON 		& 70,737 & 1.75 GiB 	& 537,934 & 2,737,873 & 4,826,400 	& 3,393,257 \\
		& Parquet 	& 1 & 4.63 MiB 	& 13 & 123 & 30 	& 161 \\
		& PDF 		& 11,442,231 & 30.48 GiB 	& 46,514,046 & 1,876,490,940 & 432,038,521 	& 18,927,559,013 \\
		& SQLite 	& 1,408 & 83.70 MiB 	& 8,839 & 186,687 & 17,973 	& 329,783 \\
		& TSV 		& 4,374 & 419.82 MiB 	& 75,989 & 569,475 & 210,506 	& 696,084 \\
		& YAML 		& 3,185 & 67.58 MiB 	& 2,236 & 31,849 & 4,601 	& 37,514 \\
		\cmidrule(lr){2-8}
		& Total 	& 231,153,272 & 742.10 GiB 	& 1,127,412,487 & 31,518,991,195 & 4,146,888,426 	& 512,095,779,088 \\
		\cmidrule(lr){1-8}
		\multirow{6}{*}{GitHub}& CSV 		& 122,091,982	& 59.86 TiB	& 5,481,784,256	& 7,390,202,751	& 36,457,207,467		& 13,912,319,966 \\
		& Excel 	& 15,787,659 & 3.02 TiB 	& 243,597,019 & 951,834,206 & 2,016,629,675 	& 5,775,869,104 \\
		& HTML 	& 199,059,080 & 630.31 GiB 	& 959,028,450 & 11,817,543,971 & 2,515,057,895 	& 173,115,916,693 \\
		& PDF 		& 40,022,516 & 79.00 GiB 	& 144,006,906 & 3,344,243,232 & 854,802,039 	& 51,385,307,211 \\
		& SQLite 	& 14,919,675 & 3.52 TiB 	& 84,554,112 & 728,970,698 & 165,104,490 	& 7,405,534,796 \\
		& TSV 		& 4,174,115 & 1.54 TiB 	& 94,845,931 & 256,836,169 & 739,989,036 	& 494,089,540 \\
		\cmidrule(lr){2-8}
		& Total 	& 396,055,027 & 68.62 TiB 	& 7,007,816,674 & 24,489,631,027 & 42,748,790,602 	& 252,089,037,310 \\
		\cmidrule(lr){1-8}
		\multirow{9}{*}{Total}& CSV 		& 122,182,649	& 59.86 TiB	& 5,484,049,755	& 7,405,833,692	& 36,469,695,577		& 13,930,304,408 \\
		& Excel 	& 15,930,671 & 3.02 TiB 	& 245,433,510 & 972,898,082 & 2,027,209,159 	& 5,836,624,241 \\
		& HTML 	& 418,456,737 & 1.30 TiB 	& 2,035,199,890 & 41,419,823,402 & 6,201,780,696 	& 666,200,940,390 \\
		& JSON 		& 70,737 & 1.75 GiB 	& 537,934 & 2,737,873 & 4,826,400 	& 3,393,257 \\
		& Parquet 	& 1 & 4.63 MiB 	& 13 & 123 & 30 	& 161 \\
		& PDF 		& 51,464,747 & 109.48 GiB 	& 190,520,952 & 5,220,734,172 & 1,286,840,560 	& 70,312,866,224 \\
		& SQLite 	& 14,921,083 & 3.52 TiB 	& 84,562,951 & 729,157,385 & 165,122,463 	& 7,405,864,579 \\
		& TSV 		& 4,178,489 & 1.54 TiB 	& 94,921,920 & 257,405,644 & 740,199,542 	& 494,785,624 \\
		& YAML 		& 3,185 & 67.58 MiB 	& 2,236 & 31,849 & 4,601 	& 37,514 \\
		\cmidrule(lr){2-8}
		& Total 	& 627,208,299 & 69.35 TiB 	& 8,135,229,161 & 56,008,622,222 & 46,895,679,028 	& 764,184,816,398 \\
		\bottomrule
	\end{tabular}
 
    \caption{\textbf{Summary statistics table,} showing counts of tables, bytes, columns, and tokens across GitHub and Common Crawl and the encountered file types.}
    \label{tab:stats}
\end{table}

We also consider token counts from metadata fields. We used \verb|tiktoken| \footnote{\url{https://github.com/openai/tiktoken}} to tokenize the \verb|ref|, space-separated column names, and \verb|context_metadata|. Because \verb|context_metadata| has nested JSON, we considered tokenizing the string of recursively-concatenated string values, instead of the serialized JSON itself (which includes JSON syntax such as commas, curly braces, and quotation marks). We compared this on a sample and found \textasciitilde10\% less token counts in the JSON vs non-JSON versions. We decided that was tolerable, so we treated \verb|context_metadata| as serialized JSON.

\subsection{Power-Law Like Distributions}

In examining TabLib, we found several metrics—including row-count, column-count, and domain-size (column-level unique-count) displaying distributions resembling power-law or Zipfian distributions, common in natural and social phenomena \citep{newman_power_2005}. Such distributions in our data suggest a few tables or columns hold most data, while the majority hold little. This pattern can significantly impact the design and evaluation of machine learning algorithms.

Power-law distributions are characterized by an exponent or Zipf's coefficient ($\alpha$ in $P(x) \propto x^{-\alpha}$), guiding the distribution's decay rate. Our comparison revealed a higher exponent in column-count than in row-count, suggesting a faster decay and affirming the typical practice of constructing tables with rows for entities and columns for entity properties (dimension tables).

Using the \verb|powerlab| library \citep{alstott_powerlaw_2014}, we observed exponents below 2 (e.g., $\alpha_{rc} \approx 1.5$ for row count), which is crucial since distributions with exponents under 2 lack well-defined mean or variance---a hallmark of true power-law distributions. Hence, the mean values of these metrics in our dataset might not accurately represent the data due to skewness from a few large tables.

Considering the distributions’ long-tail nature, training models on raw data might present challenges \citep{johnson_survey_2019}. Therefore, we propose training on aggregated tabular data instead. This approach, involving the compression of columns into concise and finite representations, could improve the robustness and generalizability of the resulting models, effectively addressing the issues posed by long-tail distributions.

An important caveat is that there may be a selection bias affecting this analysis, due to factors such as our exclusion of tables larger than 1 GB, Common Crawl's truncation of large responses, parsing bugs and limitations, etc. We leave a more detailed study of these factors for future work.

\begin{figure}[H]
	\centering
	\includegraphics[width=0.9\textwidth]{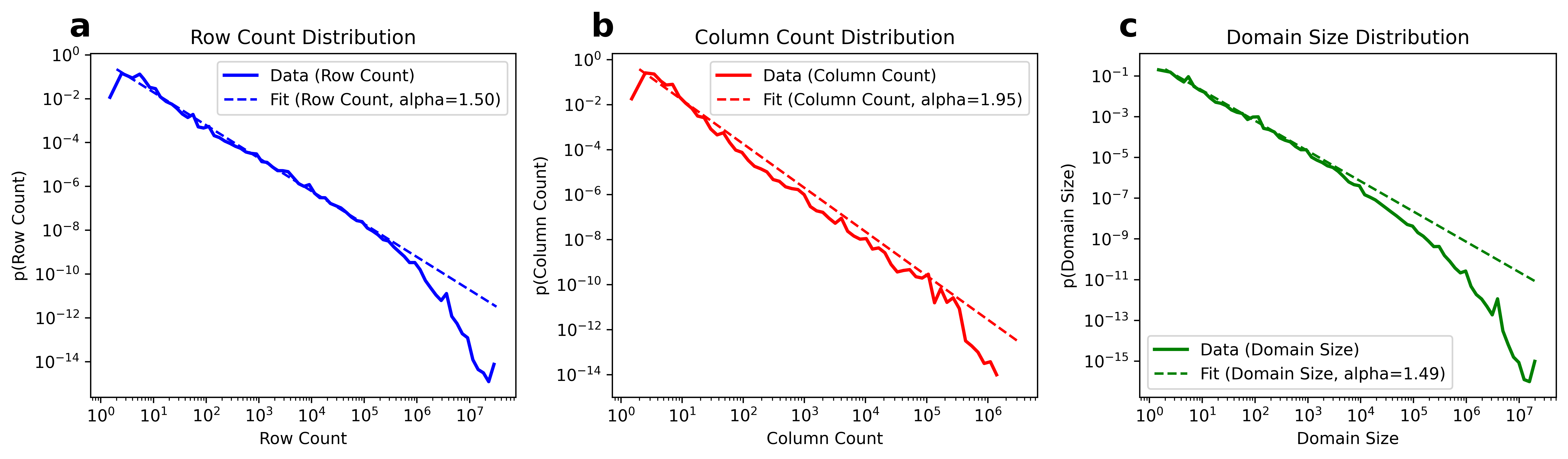}
	\caption{\textbf{Power law behavior of table statistics}. The (a) row-count, (b) column-count, and (c) domain-size (column-level unique-count) exhibit power-law-esque distributions, with a tail end following less close to a theoretical fit. The solid line shows the empirical distribution and the dotted line shows the theoretical fit given the relevant alpha value.}
    \label{fig:row_count_distribution_zipf}
\end{figure}

\subsection{Data Duplication}\label{sec:duplication}

Data duplication is a common occurrence, and is important for downstream tasks. Some works have shown that deduplication of training data can enhance language model performance \citep{lee_deduplicating_2022}, necessitating an investigation into TabLib's duplicated tables.

As seen in \autoref{tab:keys_breakdown} prior, there are many duplicates of the content hash values within the key field of the dataset. This is to be expected - many tables are duplicated across the web since they are used in different contexts by different groups of people. Within GitHub for example, there are many repositories that contain the same data, but with different names, or different versions of the same data. Whether it is an HTML table used in a frontend component, or a CSV file used in a popular data science project, there are many reasons for datasets with different contexts but the same content. Additionally, we believe that some part of this is due the practice of forking repositories on Github. See Appendix \ref{sec:duplicate-example} for examples. To look at the duplication in the dataset, we use the \verb|content_hash| of the table.

\begin{figure}[!ht]
    \centering
    \includegraphics[width=0.6\textwidth]{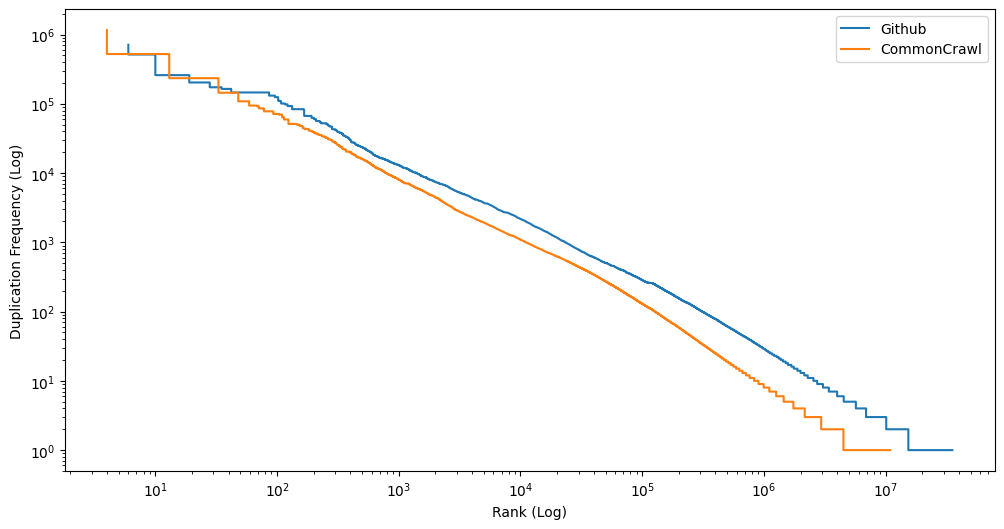}
    \caption{\textbf{Content Hash Duplication Frequencies By Source.} Duplication based on \texttt{content\_hash} shows a Zipf-like distribution when comparing frequency versus rank for both Github and Common Crawl.}
    \label{fig:content_hash_dup_freqs}
\end{figure}

In Figure \ref{fig:content_hash_dup_freqs}, we see that the behavior appears Zipf-ian, with roughly similar parameters in both sources. A notable divergence occurs around the rank 50-100 area, where GitHub has more "uneven bumps". We hypothesize that this is due to GitHub having mechanisms to copy data directly built into the platform, changing the nature of what data are commonly found. 

\begin{figure}[!ht]
    \centering
    \includegraphics[width=1.0\textwidth]{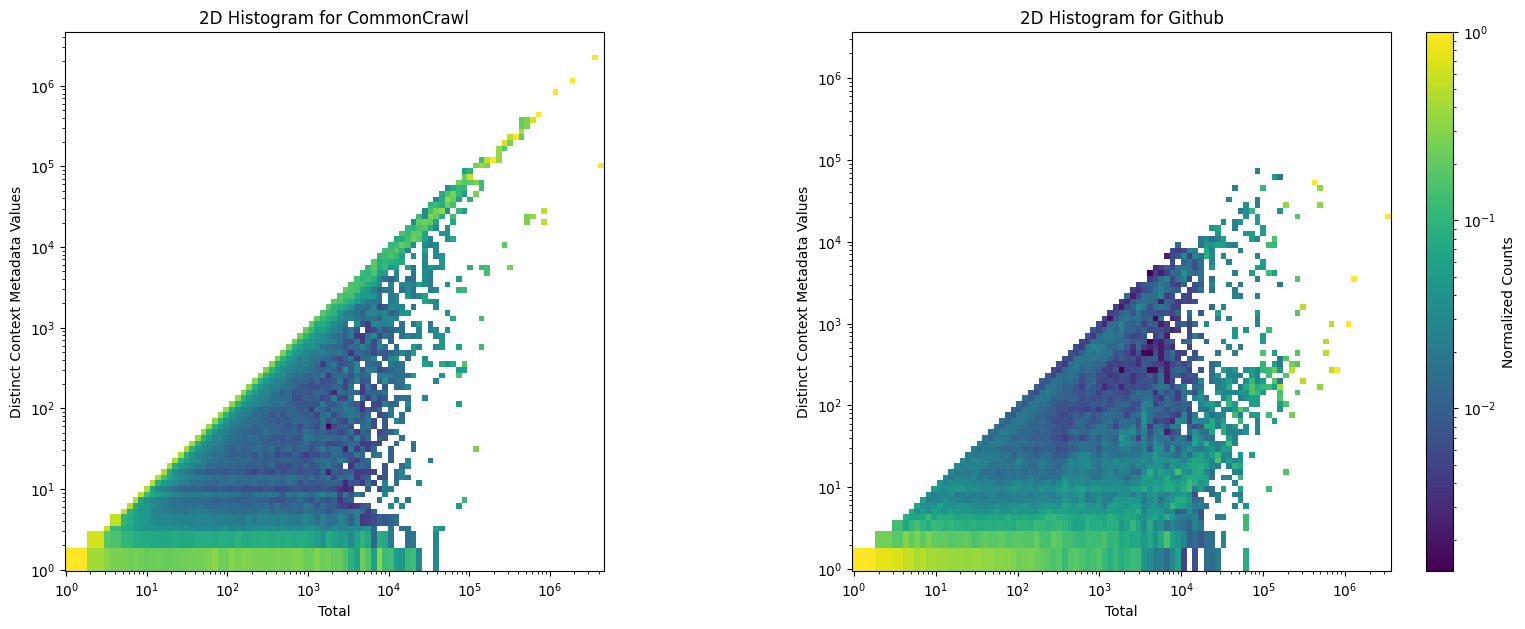}
    \caption{\textbf{2D histogram of content hash distinct values.} There is a wide variance of duplicate \texttt{context\_metadata} values among tables with duplicated \texttt{content\_hash}, for both CommonCrawl and Github. The y-axis is the log of the distinct \texttt{context\_metadata} counts, and the x-axis is the log of the total number of duplicated values for a given content hash. Both are on log scale with log bins, and the color reflects a normalized density. }
    \label{fig:2d_histogram_combined}
\end{figure}

We also consider duplicate data with different contexts. For a given set of tables with $N$ duplicate \verb|content_hash|es, there may be anywhere from $0$ to $N$ distinct  \verb|context_metadata| values for those tables. Using \verb|before| and \verb|after| fields, we compare total vs. distinct values among the duplicate content hash tables using a 2D histogram, color coded by density, shown in \autoref{fig:2d_histogram_combined}. As illustrated by the color, most of the values can be seen in the bottom left corner, which are the smaller tables. The values along the line $y=x$ have a high degree of uniqueness in \verb|context_metadata| among the same duplicated content, whereas the values along the line $y=0$ have higher degrees of duplication. There is a wide variety of data spanning those values, with high normalized counts along both $y=x$ and $y=0$, suggesting a diverse distribution of distinct \verb|context_metadata| values among tables with duplicate content hashes. We leave further investigation of the implication of filtering values along such a distribution for downstream tasks to further works.

\subsection{Data Categories}\label{sec:categories}
\begin{figure}[!ht]
    \centering
    \includegraphics[width=1.0\textwidth]{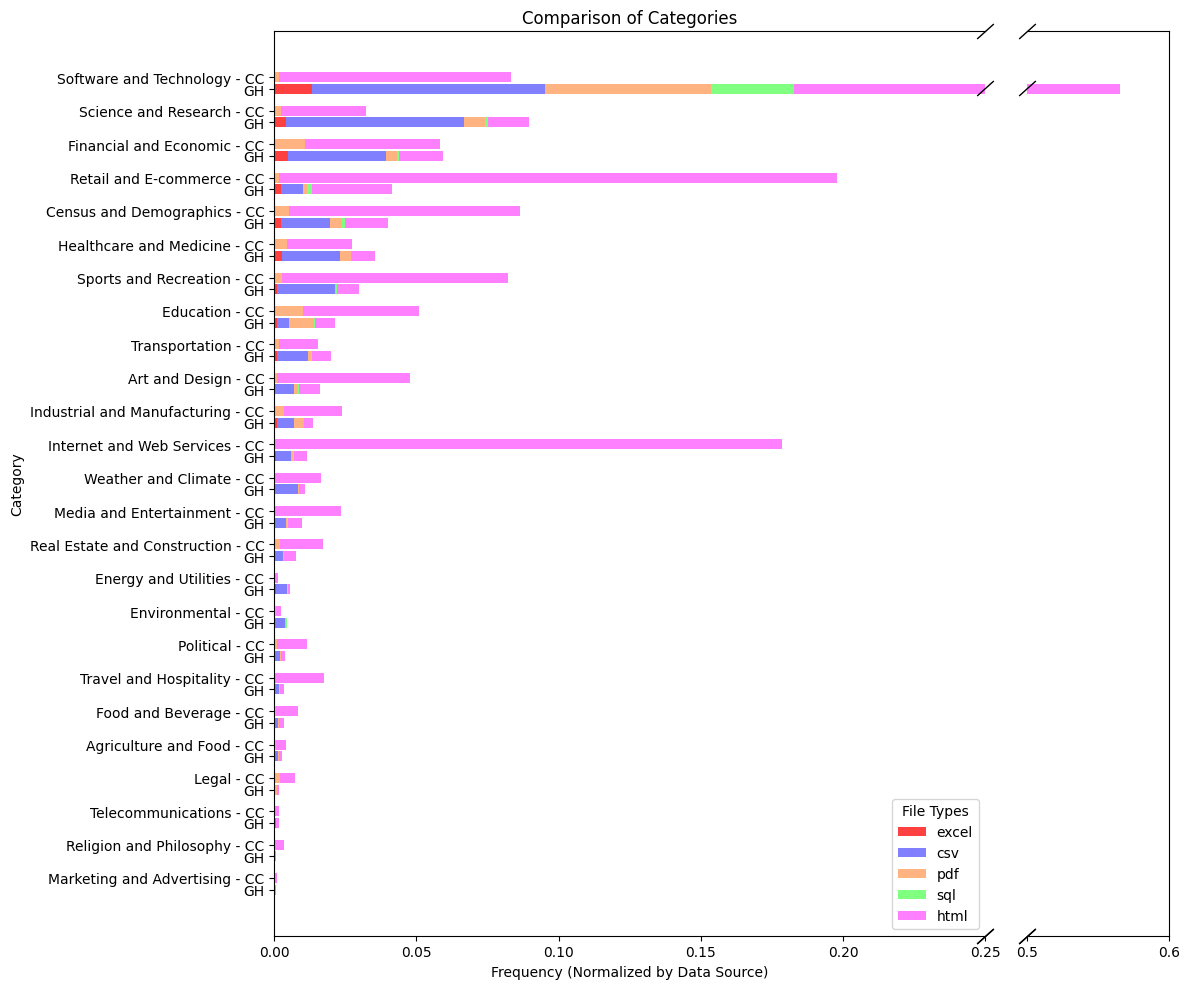}
    \caption{\textbf{Data Categories Breakdown by File Type and Data Source.} CC is Common Crawl, and GH is GitHub. HTML is the majority of content across most categories, and GitHub is predominantly of the category ``Software and Technology''. Note the x-axis has frequencies normalized by data source, and the y-axis of categories is sorted based on the normalized frequency values on GitHub. The x-axis is broken to prevent the high proportion of ``Software and Technology'' for GitHub from dominating the figure.}
    \label{fig:data_categories}
\end{figure}

There are an abundance of categories of tables in the real world, and we consider it critical to represent them in a single dataset. While TabLib includes table metadata, there are no explicit ground-truth labels for table categories such ``Sports and Recreation'' or ``Financial and Economic''. So we used the \verb|gpt-3.5-turbo| model\footnote{\url{https://platform.openai.com/docs/models/gpt-3-5}} to categorize tables using the \verb|ref| and the dataframe ``head'' (the column names and first few rows of the table), using 25 hand-picked categories. We randomly sampled 28,630 tables from TabLib and prompted \verb|gpt-3.5-turbo| to categorize them, using enums with the OpenAI function call interface\footnote{\url{https://platform.openai.com/docs/plugins/getting-started/writing-descriptions}}. We discarded 2,364 responses which did not exactly match a requested enum value. The results of this categorization are shown in \autoref{fig:data_categories}.

As shown in \autoref{fig:data_categories}, the majority of GitHub tables are centered around the category ``Software and Technology'' which includes many examples of code and documentation. Outside of code-related content, there are a variety of content types including: science and research, financial and economic, retail and e-commerce, etc. Common Crawl is more balanced and diverse, with a majority of tables focused on retail and e-commerce, internet and web services, and calendars, etc. Most of the Common Crawl tables were HTML, whereas in GitHub most those HTML tables occurred in the ``Software and Technology'' category as documentation. 

Having the categories may be useful for downstream tasks, such as training a model to classify or generate tables of a specific category \citep{korini_column_2023}. We chose a limited set of categories to label and example tables to process, and leave further investigation of the accuracy and effectiveness of these categories to future work.

\subsection{Language Breakdown}
Another important aspect of diversity for language models is the language itself, as discussed in many papers such as LAION-5B \citep{schuhmann_laion-5b_2022} and the Pile \citep{gao_pile_2020}. We classified the language of tables using \verb|langdetect|\footnote{\url{https://github.com/Mimino666/langdetect}}, \verb|fasttext|\footnote{\url{https://github.com/facebookresearch/fastText}}, and \verb|gpt-3.5-turbo|, based on the column names and values of string-typed cells, joined by spaces and limited to 100 characters. With manual inspection on a small sample, \verb|gpt-3.5-turbo| was the most accurate. 

We sampled 10,000 random tables from TabLib and classified their languages using \verb|gpt-3.5-turbo|, with results shown in \ref{fig:lang_counts}. Since English was 69\% of the data, English is excluded from the figure. A large portion of tables were classified as ``Unknown'', which includes mostly numeric tables which include no human languages. See \nameref{sec:unknown-lang-example} in the appendix for an example of a table with an ``Unknown'' language.

\begin{figure}[!ht]
\begin{tikzpicture}
\begin{axis}[
    axis lines=left,
    bar width=4pt,
    xtick=data,
    enlarge x limits=0.015,
    symbolic x coords={Spanish,Russian,Japanese,Chinese,Unknown,German,French,Italian,Korean,Polish,Portuguese,Dutch,Czech,Ukrainian,Turkish,Vietnamese,Romanian,Indonesian,Arabic,Greek,Swedish,Finnish,Slovak,Danish,Norwegian,Persian,Hungarian,Thai,Bulgarian,Slovenian,Catalan,Hindi,Lithuanian,Bengali,Croatian,Tamil,Serbian,Albanian,Latvian,Estonian,Hebrew,Latin,Urdu,Nepali,Bahasa Indonesia,Basque,Welsh,Irish,Haitian Creole,Georgian,Malayalam,Kurdish,Uzbek,Malay,Punjabi,Kannada,Somali,Telugu,Marathi,Sinhala,Khmer,Sanskrit,Swahili,Icelandic,Tagalog,Pashto,Macedonian,Luxembourgish,Odia,Bosnian},
    x tick label style={
        /pgf/number format/1000 sep=,
        rotate=45,
        anchor=east,
        font=\tiny
    },
    ybar,
    enlarge y limits={upper,value=0.05},
    ymin=0,
    ytick={0,0.5,...,3},
    ylabel={\% Frequency},
    width=\textwidth,
    height=200,
]
\addplot 
    table[col sep=comma, x=Language,y=Percentage] {images/lang-props.csv};
\end{axis}
\end{tikzpicture}
\caption{\textbf{Frequency estimate of non-English languages in TabLib.} Note that English had a frequency of 69\% so was excluded from this figure. All languages shown had a non-zero frequency in the 10,000 table sample.}
\label{fig:lang_counts}
\end{figure}


\subsection{Data Types Breakdown}
In addition to language, tabular data also has a variety of column types. We look at the type breakdown of the columns in the dataset. \autoref{tab:dtypes_breakdown} below shows the column type frequency based on the inferred table schema. Surprisingly, we found that very little of the data had timestamp or datetime columns. This is likely due to implementation details of Pandas' type inference, requiring a separate pass to parse dates and timestamps in their various forms. In some cases it may be difficult or impossible to correctly infer timestamps, such as integral UNIX epoch timestamps. We believe that overall, the distribution is dominated by parsing decisions since the data in many formats (HTML, CSV, TSV) are stored as strings first, and column type is then inferred. We leave more detailed data cleaning, post-processing, and type inference to future works.

\begin{table}[!ht]
    \centering
    \begin{tabular}{ll}
        \toprule
        Data Type 	& \% of Tables \\
        \midrule
        String 		& 61.8\% \\
        Float 		& 22.3\% \\
        Integer 	& 11.3\% \\	
        Unknown 	& 3.8\% \\
        Boolean 	& 0.74\% \\
        Timestamp 	& 0.04\% \\
        Datetime 	& 0.01\% \\
        \bottomrule
    \end{tabular}
    \caption{\textbf{Column type frequency.} The majority of column types are strings.}
    \label{tab:dtypes_breakdown}
\end{table}

\subsection{Embeddings}
\begin{figure}[!ht]
    \centering
    \fbox{\includegraphics[width=0.6\textwidth]{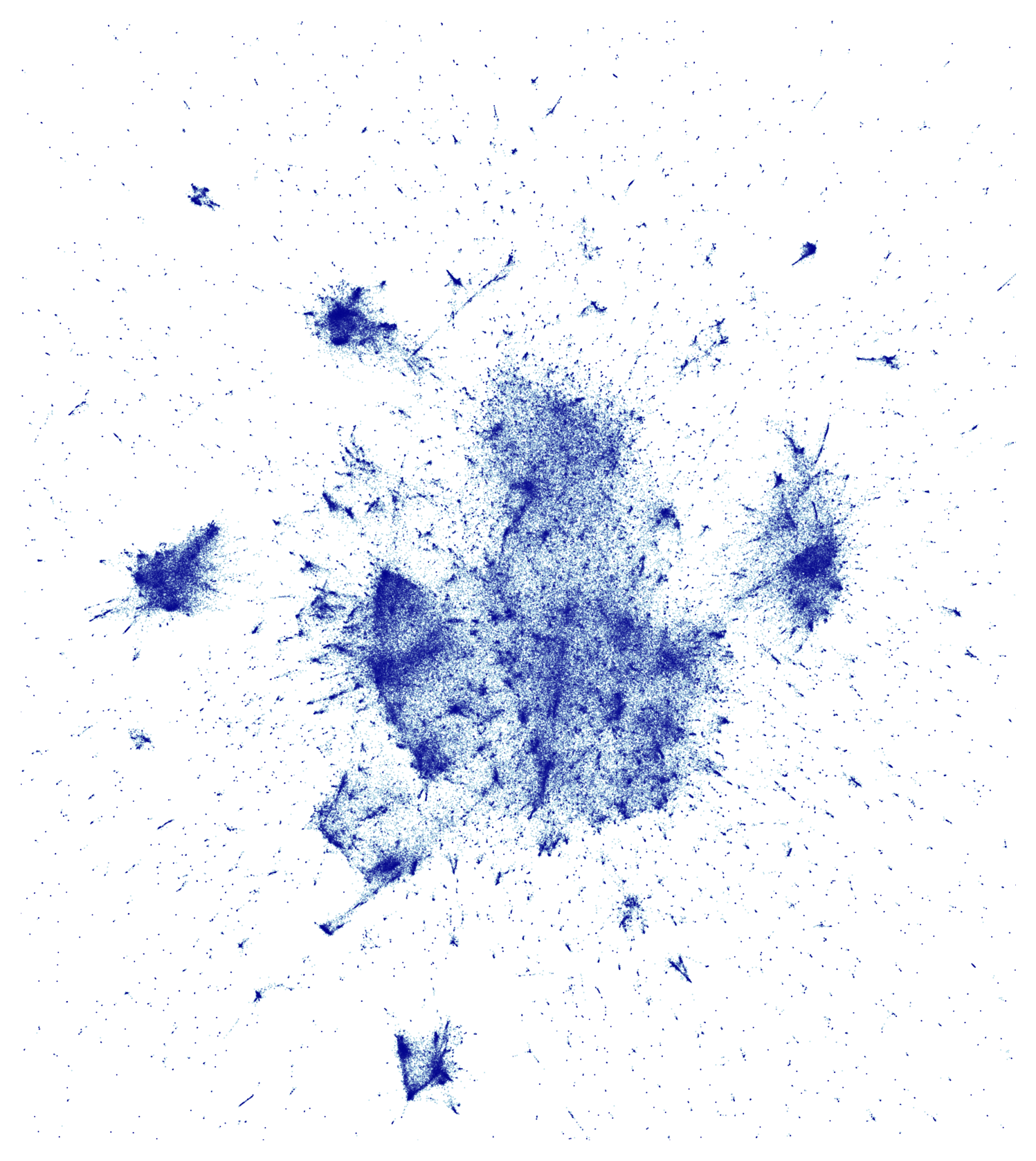}}
    \caption{\textbf{UMAP sample.} A UMAP embedding plot generated from a sample of 500K tables from TabLib, using column names and the first few rows from each table.}
    \label{fig:umap_sample}
\end{figure}

Word embeddings are vector representations of words that contain semantic meaning \citep{mikolov_efficient_2013}. We can represent other features such as column names, table schemas, etc. using these word embeddings. We sampled 500,000 tables from TabLib and used the \verb|all-MiniLM-L6-v2| model of Sentence Transformers \citep{reimers_sentence-bert_2019} to embed the column names and first few rows of each table into a word embedding. We then computed a UMAP embedding to project those into a 2D plot, shown in \autoref{fig:umap_sample}. As we can see, there are many large and small clusters. Upon manual inspection, the large clusters tend to represent different languages, and the smaller clusters align semantically towards categories (see \nameref{sec:categories}). This technique focuses mainly on table metadata such as column names and schema, and does a poor job of representing the contents of the table itself, which we leave for future work.

\section{Discussion}

\subsection{Ethics}

\subsubsection{Personally Identifiable Information}

TabLib captures personally identifiable information (PII), such as names, phone numbers, and email addresses. However, all data within TabLib are from publicly accessible sources, implying that the PII it contains is already available to the public. Furthermore, we acknowledge that the identification and protection of PII is an evolving field of study, and we believe that raw datasets like TabLib will be essential resources in this research.

\subsubsection{Potential Biases}

Publicly available data, like the data in TabLib, often contain inherent biases which can be inadvertently propagated in trained models. This phenomena, well-documented in language and image models, might also permeate tabular data, whether within the actual tabular data or their accompanying descriptive context. Acknowledging the presence of possible biases, TabLib presents an opportunity to study and mitigate such prejudices, leading to the development of fairer AI systems.

\subsubsection{Legality of Content}

The legal implications of training machine learning models using copyrighted data is a topic of ongoing debate within the machine learning community \citep{gao_pile_2020}. However, there is much less discussion, and even less clarity on the processing and distribution of data for research purposes. Based on our understanding, we believe this falls under the purview of fair use. Additionally, it is noteworthy to mention that under U.S. copyright law, facts and data are not subject to copyright protection (see \emph{Feist v. Rural Telephone}\footnote{\url{https://www.law.cornell.edu/supremecourt/text/499/340}}). This aspect of the law, while not providing definitive legal clarity, adds an interesting dimension to the discussion surrounding the use of datasets like TabLib, which collect factual data in tables. We commit to remaining informed and making necessary adjustments as the legal implications of this work become clearer.

\subsubsection{Data Licensing}

TabLib is an aggregation of publicly available data. Each datum has its own specific license which must be respected. We have attempted to include provenance information for each table within its \verb|context_metadata| to help find licensing information. We also recommend that this dataset be used primarily for research purposes.

\subsection{Limitations}

\subsubsection{Source Limitations}
TabLib's initial version does not include many public sources such as CKAN sources (e.g., \href{https://data.gov}{data.gov} and \href{https://data.gov.uk}{data.gov.uk}), books (e.g., \href{https://www.gutenberg.org/}{Project Gutenberg}), and other datasets on the public Internet not indexed by Common Crawl. Additionally, we have not included source files larger than 1 GB, GitHub branches other than "main" or "master", or truncated Common Crawl responses. These limitations affect the diversity, volume, and distribution of data in TabLib.

\subsubsection{Parsing Limitations}
Detecting and parsing table structures is difficult, and our current parsing capabilities are limited. For instance, PDF tables that span multiple pages are not recognized as a single table. Similarly, ambiguities in the meaning of ``before'' and ``after'' can result in PDF tables with missing or incomplete context. For HTML tables, the presence of JavaScript, CSS, and other elements can introduce noise into the context. Furthermore, our current version does not support the extraction of tables from images, whether they are standalone image files or inlined in PDFs and HTML.  Another challenge lies in the accurate inference of column types and the correct detection of column headers (e.g. nested column headers). These limitations could potentially affect the accuracy of the data extracted and its subsequent usability.

\subsubsection{Metadata Limitations}
Metadata are often inaccurate, incomplete, or missing. This includes data we actively sought to include, such as provenance. It also includes data that are useful but were not intentionally captured, such as licensing. 

\section{Future Work}
There are numerous areas for exploration and improvement to enhance the value of TabLib as a research asset.
\begin{itemize}
    \item \textbf{Add New Data Sources:} Increase the size of TabLib by including other Common Crawl crawls, GitHub branches beyond master and main, and broader expansion beyond the limitations of Common Crawl.
    \item \textbf{Derive New Tables:} Programmatically transform existing data tables to create new data tables, thereby increasing the number of tables.
    \item \textbf{Enhanced Table Extraction:} Improve our current table extraction methods, particularly for complex formats like PDFs and images, to increase the accuracy and completeness of the data extracted.
    \item \textbf{Inclusion of Additional Metadata:} Include additional metadata, such as licensing, categorization, etc.
    \item \textbf{Creation of Cleaned Versions:} Develop cleaned versions of TabLib by removing categories of information such as noise, PII, etc., thereby increasing the usability of the dataset for various applications.
    \item \textbf{Development of Benchmarks:} Create benchmarks around TabLib for tasks like question answering and search, to encourage the use of this dataset and spur advancements in tabular data research.
    \item \textbf{Pre-training Large Data Models:} Explore the potential of pre-training large data models exclusively on TabLib’s tabular data.
    \item \textbf{Bias Study and Mitigation:} Study social biases in tabular data and develop techniques to mitigate them.
\end{itemize}

\section{Conclusion}
\subsection{Key Outcomes}

In this work, we present TabLib, a dataset of 627 million tables (69 TiB) with 867 billion tokens of context extracted from GitHub and Common Crawl. TabLib contains raw, minimally processed tabular data derived from formats like CSV, HTML, PDF, and Excel, along with rich contextual metadata.

Our analysis of TabLib shows its extensive coverage across a multitude of topics, languages, and data types. The dataset exhibits interesting long-tail behavior with important consequences for downstream training and evaluation. Furthermore, our duplication analysis confirms that a non-trivial portion of TabLib consists of unique tables, and a large majority of tables contains unique metadata, further enhancing its value as a resource for AI research and development.

\subsection{Acknowledgements}

We would like to thank TPU Research Cloud\footnote{\url{https://sites.research.google/trc}} for providing the compute resources to process the data.

We are grateful to GitHub and Common Crawl for making available the underlying data necessary for TabLib.

\clearpage

\bibliographystyle{unsrtnat}
\bibliography{references}  

\begin{thebibliography}{59}
\providecommand{\natexlab}[1]{#1}
\providecommand{\url}[1]{\texttt{#1}}
\expandafter\ifx\csname urlstyle\endcsname\relax
  \providecommand{\doi}[1]{doi: #1}\else
  \providecommand{\doi}{doi: \begingroup \urlstyle{rm}\Url}\fi

\bibitem[Hoffmann et~al.(2022)Hoffmann, Borgeaud, Mensch, Buchatskaya, Cai, Rutherford, Casas, Hendricks, Welbl, Clark, Hennigan, Noland, Millican, Driessche, Damoc, Guy, Osindero, Simonyan, Elsen, Rae, Vinyals, and Sifre]{hoffmann_training_2022}
Jordan Hoffmann, Sebastian Borgeaud, Arthur Mensch, Elena Buchatskaya, Trevor Cai, Eliza Rutherford, Diego de~Las Casas, Lisa~Anne Hendricks, Johannes Welbl, Aidan Clark, Tom Hennigan, Eric Noland, Katie Millican, George van~den Driessche, Bogdan Damoc, Aurelia Guy, Simon Osindero, Karen Simonyan, Erich Elsen, Jack~W. Rae, Oriol Vinyals, and Laurent Sifre.
\newblock Training {Compute}-{Optimal} {Large} {Language} {Models}, March 2022.
\newblock URL \url{http://arxiv.org/abs/2203.15556}.
\newblock arXiv:2203.15556 [cs].

\bibitem[Zha et~al.(2023{\natexlab{a}})Zha, Bhat, Lai, Yang, Jiang, Zhong, and Hu]{zha_data-centric_2023}
Daochen Zha, Zaid~Pervaiz Bhat, Kwei-Herng Lai, Fan Yang, Zhimeng Jiang, Shaochen Zhong, and Xia Hu.
\newblock Data-centric {Artificial} {Intelligence}: {A} {Survey}, June 2023{\natexlab{a}}.
\newblock URL \url{http://arxiv.org/abs/2303.10158}.
\newblock Issue: arXiv:2303.10158 arXiv:2303.10158 [cs].

\bibitem[Radford et~al.(2021)Radford, Kim, Hallacy, Ramesh, Goh, Agarwal, Sastry, Askell, Mishkin, Clark, Krueger, and Sutskever]{radford_learning_2021}
Alec Radford, Jong~Wook Kim, Chris Hallacy, Aditya Ramesh, Gabriel Goh, Sandhini Agarwal, Girish Sastry, Amanda Askell, Pamela Mishkin, Jack Clark, Gretchen Krueger, and Ilya Sutskever.
\newblock Learning {Transferable} {Visual} {Models} {From} {Natural} {Language} {Supervision}, February 2021.
\newblock URL \url{http://arxiv.org/abs/2103.00020}.
\newblock arXiv:2103.00020 [cs].

\bibitem[Ramesh et~al.(2021)Ramesh, Pavlov, Goh, Gray, Voss, Radford, Chen, and Sutskever]{ramesh_zero-shot_2021}
Aditya Ramesh, Mikhail Pavlov, Gabriel Goh, Scott Gray, Chelsea Voss, Alec Radford, Mark Chen, and Ilya Sutskever.
\newblock Zero-{Shot} {Text}-to-{Image} {Generation}, February 2021.
\newblock URL \url{http://arxiv.org/abs/2102.12092}.
\newblock arXiv:2102.12092 [cs].

\bibitem[Schuhmann et~al.(2021)Schuhmann, Vencu, Beaumont, Kaczmarczyk, Mullis, Katta, Coombes, Jitsev, and Komatsuzaki]{schuhmann_laion-400m_2021}
Christoph Schuhmann, Richard Vencu, Romain Beaumont, Robert Kaczmarczyk, Clayton Mullis, Aarush Katta, Theo Coombes, Jenia Jitsev, and Aran Komatsuzaki.
\newblock {LAION}-{400M}: {Open} {Dataset} of {CLIP}-{Filtered} 400 {Million} {Image}-{Text} {Pairs}, November 2021.
\newblock URL \url{http://arxiv.org/abs/2111.02114}.
\newblock Issue: arXiv:2111.02114 arXiv:2111.02114 [cs].

\bibitem[Schuhmann et~al.(2022)Schuhmann, Beaumont, Vencu, Gordon, Wightman, Cherti, Coombes, Katta, Mullis, Wortsman, Schramowski, Kundurthy, Crowson, Schmidt, Kaczmarczyk, and Jitsev]{schuhmann_laion-5b_2022}
Christoph Schuhmann, Romain Beaumont, Richard Vencu, Cade Gordon, Ross Wightman, Mehdi Cherti, Theo Coombes, Aarush Katta, Clayton Mullis, Mitchell Wortsman, Patrick Schramowski, Srivatsa Kundurthy, Katherine Crowson, Ludwig Schmidt, Robert Kaczmarczyk, and Jenia Jitsev.
\newblock {LAION}-{5B}: {An} open large-scale dataset for training next generation image-text models, October 2022.
\newblock URL \url{http://arxiv.org/abs/2210.08402}.
\newblock Issue: arXiv:2210.08402 arXiv:2210.08402 [cs].

\bibitem[Rombach et~al.(2022)Rombach, Blattmann, Lorenz, Esser, and Ommer]{rombach_high-resolution_2022}
Robin Rombach, Andreas Blattmann, Dominik Lorenz, Patrick Esser, and Björn Ommer.
\newblock High-{Resolution} {Image} {Synthesis} with {Latent} {Diffusion} {Models}, April 2022.
\newblock URL \url{http://arxiv.org/abs/2112.10752}.
\newblock arXiv:2112.10752 [cs].

\bibitem[Badaro et~al.(2023)Badaro, Saeed, and Papotti]{badaro_transformers_2023}
Gilbert Badaro, Mohammed Saeed, and Paolo Papotti.
\newblock Transformers for {Tabular} {Data} {Representation}: {A} {Survey} of {Models} and {Applications}.
\newblock \emph{Transactions of the Association for Computational Linguistics}, 11:\penalty0 227--249, March 2023.
\newblock ISSN 2307-387X.
\newblock \doi{10.1162/tacl_a_00544}.
\newblock URL \url{https://direct.mit.edu/tacl/article/doi/10.1162/tacl_a_00544/115239/Transformers-for-Tabular-Data-Representation-A}.

\bibitem[Jin et~al.(2022)Jin, Siebert, Li, and Chen]{jin_survey_2022}
Nengzheng Jin, Joanna Siebert, Dongfang Li, and Qingcai Chen.
\newblock A {Survey} on {Table} {Question} {Answering}: {Recent} {Advances}, July 2022.
\newblock URL \url{http://arxiv.org/abs/2207.05270}.
\newblock arXiv:2207.05270 [cs].

\bibitem[Dong et~al.(2022)Dong, Cheng, He, Zhou, Zhou, Zhou, Liu, Han, and Zhang]{dong_table_2022}
Haoyu Dong, Zhoujun Cheng, Xinyi He, Mengyu Zhou, Anda Zhou, Fan Zhou, Ao~Liu, Shi Han, and Dongmei Zhang.
\newblock Table {Pre}-training: {A} {Survey} on {Model} {Architectures}, {Pre}-training {Objectives}, and {Downstream} {Tasks}, April 2022.
\newblock URL \url{http://arxiv.org/abs/2201.09745}.
\newblock arXiv:2201.09745 [cs].

\bibitem[Lehmberg et~al.(2016)Lehmberg, Ritze, Meusel, and Bizer]{lehmberg_large_2016}
Oliver Lehmberg, Dominique Ritze, Robert Meusel, and Christian Bizer.
\newblock A {Large} {Public} {Corpus} of {Web} {Tables} containing {Time} and {Context} {Metadata}.
\newblock In \emph{Proceedings of the 25th {International} {Conference} {Companion} on {World} {Wide} {Web} - {WWW} '16 {Companion}}, pages 75--76, Montr\&\#233;al, Qu\&\#233;bec, Canada, 2016. ACM Press.
\newblock ISBN 978-1-4503-4144-8.
\newblock \doi{10.1145/2872518.2889386}.
\newblock URL \url{http://dl.acm.org/citation.cfm?doid=2872518.2889386}.

\bibitem[Bhagavatula et~al.(2015)Bhagavatula, Noraset, and Downey]{arenas_tabel_2015}
Chandra~Sekhar Bhagavatula, Thanapon Noraset, and Doug Downey.
\newblock {TabEL}: {Entity} {Linking} in {Web} {Tables}.
\newblock In Marcelo Arenas, Oscar Corcho, Elena Simperl, Markus Strohmaier, Mathieu d'Aquin, Kavitha Srinivas, Paul Groth, Michel Dumontier, Jeff Heflin, Krishnaprasad Thirunarayan, Krishnaprasad Thirunarayan, and Steffen Staab, editors, \emph{The {Semantic} {Web} - {ISWC} 2015}, volume 9366, pages 425--441. Springer International Publishing, Cham, 2015.
\newblock ISBN 978-3-319-25006-9 978-3-319-25007-6.
\newblock \doi{10.1007/978-3-319-25007-6_25}.
\newblock URL \url{http://link.springer.com/10.1007/978-3-319-25007-6_25}.
\newblock Series Title: Lecture Notes in Computer Science.

\bibitem[Hulsebos et~al.(2023)Hulsebos, Demiralp, and Groth]{hulsebos_gittables_2023}
Madelon Hulsebos, Çağatay Demiralp, and Paul Groth.
\newblock {GitTables}: {A} {Large}-{Scale} {Corpus} of {Relational} {Tables}.
\newblock \emph{Proceedings of the ACM on Management of Data}, 1\penalty0 (1):\penalty0 1--17, May 2023.
\newblock ISSN 2836-6573.
\newblock \doi{10.1145/3588710}.
\newblock URL \url{http://arxiv.org/abs/2106.07258}.
\newblock arXiv:2106.07258 [cs].

\bibitem[Hu et~al.(2019)Hu, Gaikwad, Bakker, Hulsebos, Zgraggen, Hidalgo, Kraska, Li, Satyanarayan, and Demiralp]{hu_viznet_2019}
Kevin Hu, Neil Gaikwad, Michiel Bakker, Madelon Hulsebos, Emanuel Zgraggen, César Hidalgo, Tim Kraska, Guoliang Li, Arvind Satyanarayan, and Çağatay Demiralp.
\newblock {VizNet}: {Towards} {A} {Large}-{Scale} {Visualization} {Learning} and {Benchmarking} {Repository}, May 2019.
\newblock URL \url{http://arxiv.org/abs/1905.04616}.
\newblock arXiv:1905.04616 [cs].

\bibitem[Vogel and Binnig(2023)]{vogel_wikidbs_2023}
Liane Vogel and Carsten Binnig.
\newblock {WikiDBs}: {A} {Corpus} of {Relational} {Databases} {From} {Wikidata}.
\newblock In \emph{Joint {Proceedings} of {Workshops} at the 49th {International} {Conference} on {Very} {Large} {Data} {Bases} ({VLDB} 2023), {Vancouver}, {Canada}, {August} 28 - {September} 1, 2023}, volume 3462 of \emph{{CEUR} {Workshop} {Proceedings}}. CEUR-WS.org, 2023.
\newblock URL \url{https://ceur-ws.org/Vol-3462/TADA3.pdf}.

\bibitem[Yu et~al.(2019)Yu, Zhang, Yang, Yasunaga, Wang, Li, Ma, Li, Yao, Roman, Zhang, and Radev]{yu_spider_2019}
Tao Yu, Rui Zhang, Kai Yang, Michihiro Yasunaga, Dongxu Wang, Zifan Li, James Ma, Irene Li, Qingning Yao, Shanelle Roman, Zilin Zhang, and Dragomir Radev.
\newblock Spider: {A} {Large}-{Scale} {Human}-{Labeled} {Dataset} for {Complex} and {Cross}-{Domain} {Semantic} {Parsing} and {Text}-to-{SQL} {Task}, February 2019.
\newblock URL \url{http://arxiv.org/abs/1809.08887}.
\newblock arXiv:1809.08887 [cs].

\bibitem[Cafarella et~al.(2008)Cafarella, Halevy, Wang, Wu, and Zhang]{cafarella_webtables_2008}
Michael~J. Cafarella, Alon Halevy, Daisy~Zhe Wang, Eugene Wu, and Yang Zhang.
\newblock {WebTables}: exploring the power of tables on the web.
\newblock \emph{Proceedings of the VLDB Endowment}, 1\penalty0 (1):\penalty0 538--549, August 2008.
\newblock ISSN 2150-8097.
\newblock \doi{10.14778/1453856.1453916}.
\newblock URL \url{https://doi.org/10.14778/1453856.1453916}.

\bibitem[Benjelloun et~al.(2020)Benjelloun, Chen, and Noy]{benjelloun_google_2020}
Omar Benjelloun, Shiyu Chen, and Natasha Noy.
\newblock Google {Dataset} {Search} by the {Numbers}, June 2020.
\newblock URL \url{http://arxiv.org/abs/2006.06894}.
\newblock arXiv:2006.06894 [cs].

\bibitem[Chapman et~al.(2020)Chapman, Simperl, Koesten, Konstantinidis, Ibáñez-Gonzalez, Kacprzak, and Groth]{chapman_dataset_2020}
Adriane Chapman, Elena Simperl, Laura Koesten, George Konstantinidis, Luis-Daniel Ibáñez-Gonzalez, Emilia Kacprzak, and Paul Groth.
\newblock Dataset search: a survey.
\newblock \emph{The VLDB Journal}, 29\penalty0 (1):\penalty0 251--272, January 2020.
\newblock ISSN 1066-8888, 0949-877X.
\newblock \doi{10.1007/s00778-019-00564-x}.
\newblock URL \url{http://arxiv.org/abs/1901.00735}.
\newblock Number: 1 arXiv:1901.00735 [cs].

\bibitem[Zhang and Balog(2018)]{zhang_ad_2018}
Shuo Zhang and Krisztian Balog.
\newblock Ad {Hoc} {Table} {Retrieval} using {Semantic} {Similarity}.
\newblock In \emph{Proceedings of the 2018 {World} {Wide} {Web} {Conference} on {World} {Wide} {Web} - {WWW} '18}, pages 1553--1562, 2018.
\newblock \doi{10.1145/3178876.3186067}.
\newblock URL \url{http://arxiv.org/abs/1802.06159}.
\newblock arXiv:1802.06159 [cs].

\bibitem[Dong et~al.(2014)Dong, Gabrilovich, Heitz, Horn, Lao, Murphy, Strohmann, Sun, and Zhang]{dong_knowledge_2014}
Xin Dong, Evgeniy Gabrilovich, Geremy Heitz, Wilko Horn, Ni~Lao, Kevin Murphy, Thomas Strohmann, Shaohua Sun, and Wei Zhang.
\newblock Knowledge vault: a web-scale approach to probabilistic knowledge fusion.
\newblock In \emph{Proceedings of the 20th {ACM} {SIGKDD} international conference on {Knowledge} discovery and data mining}, pages 601--610, New York New York USA, August 2014. ACM.
\newblock ISBN 978-1-4503-2956-9.
\newblock \doi{10.1145/2623330.2623623}.
\newblock URL \url{https://dl.acm.org/doi/10.1145/2623330.2623623}.

\bibitem[Liu et~al.(2023)Liu, Chabot, Troncy, Huynh, Labbé, and Monnin]{liu_tabular_2023}
Jixiong Liu, Yoan Chabot, Raphaël Troncy, Viet-Phi Huynh, Thomas Labbé, and Pierre Monnin.
\newblock From tabular data to knowledge graphs: {A} survey of semantic table interpretation tasks and methods.
\newblock \emph{Journal of Web Semantics}, 76:\penalty0 100761, April 2023.
\newblock ISSN 1570-8268.
\newblock \doi{10.1016/j.websem.2022.100761}.
\newblock URL \url{https://www.sciencedirect.com/science/article/pii/S1570826822000452}.

\bibitem[Jiménez-Ruiz et~al.(2020)Jiménez-Ruiz, Hassanzadeh, Efthymiou, Chen, and Srinivas]{jimenez-ruiz_semtab_2020}
Ernesto Jiménez-Ruiz, Oktie Hassanzadeh, Vasilis Efthymiou, Jiaoyan Chen, and Kavitha Srinivas.
\newblock {SemTab} 2019: {Resources} to {Benchmark} {Tabular} {Data} to {Knowledge} {Graph} {Matching} {Systems}.
\newblock In Andreas Harth, Sabrina Kirrane, Axel-Cyrille Ngonga~Ngomo, Heiko Paulheim, Anisa Rula, Anna~Lisa Gentile, Peter Haase, and Michael Cochez, editors, \emph{The {Semantic} {Web}}, Lecture {Notes} in {Computer} {Science}, pages 514--530, Cham, 2020. Springer International Publishing.
\newblock ISBN 978-3-030-49461-2.
\newblock \doi{10.1007/978-3-030-49461-2_30}.

\bibitem[Efthymiou et~al.(2017)Efthymiou, Hassanzadeh, Rodriguez-Muro, and Christophides]{efthymiou_matching_2017}
Vasilis Efthymiou, Oktie Hassanzadeh, Mariano Rodriguez-Muro, and Vassilis Christophides.
\newblock Matching {Web} {Tables} with {Knowledge} {Base} {Entities}: {From} {Entity} {Lookups} to {Entity} {Embeddings}.
\newblock In Claudia d'Amato, Miriam Fernandez, Valentina Tamma, Freddy Lecue, Philippe Cudré-Mauroux, Juan Sequeda, Christoph Lange, and Jeff Heflin, editors, \emph{The {Semantic} {Web} – {ISWC} 2017}, Lecture {Notes} in {Computer} {Science}, pages 260--277, Cham, 2017. Springer International Publishing.
\newblock ISBN 978-3-319-68288-4.
\newblock \doi{10.1007/978-3-319-68288-4_16}.

\bibitem[Bonfitto(2021)]{bonfitto_table_2021}
Sara Bonfitto.
\newblock Table understanding approaches for extracting knowledge from heterogeneous tables, March 2021.
\newblock URL \url{https://wires.onlinelibrary.wiley.com/doi/abs/10.1002/widm.1407}.

\bibitem[Hulsebos et~al.(2019)Hulsebos, Hu, Bakker, Zgraggen, Satyanarayan, Kraska, Demiralp, and Hidalgo]{hulsebos_sherlock_2019}
Madelon Hulsebos, Kevin Hu, Michiel Bakker, Emanuel Zgraggen, Arvind Satyanarayan, Tim Kraska, Çağatay Demiralp, and César Hidalgo.
\newblock Sherlock: {A} {Deep} {Learning} {Approach} to {Semantic} {Data} {Type} {Detection}, May 2019.
\newblock URL \url{http://arxiv.org/abs/1905.10688}.
\newblock arXiv:1905.10688 [cs, stat].

\bibitem[Dong et~al.(2021)Dong, Takeoka, Xiao, and Oyamada]{dong_efficient_2021}
Yuyang Dong, Kunihiro Takeoka, Chuan Xiao, and Masafumi Oyamada.
\newblock Efficient {Joinable} {Table} {Discovery} in {Data} {Lakes}: {A} {High}-{Dimensional} {Similarity}-{Based} {Approach}, March 2021.
\newblock URL \url{http://arxiv.org/abs/2010.13273}.
\newblock arXiv:2010.13273 [cs].

\bibitem[Zhang and Balog(2019)]{zhang_recommending_2019}
Shuo Zhang and Krisztian Balog.
\newblock Recommending {Related} {Tables}, July 2019.
\newblock URL \url{http://arxiv.org/abs/1907.03595}.
\newblock arXiv:1907.03595 [cs].

\bibitem[Zhu et~al.(2019)Zhu, Deng, Nargesian, and Miller]{zhu_josie_2019}
Erkang Zhu, Dong Deng, Fatemeh Nargesian, and Renée~J. Miller.
\newblock {JOSIE}: {Overlap} {Set} {Similarity} {Search} for {Finding} {Joinable} {Tables} in {Data} {Lakes}.
\newblock In \emph{Proceedings of the 2019 {International} {Conference} on {Management} of {Data}}, {SIGMOD} '19, pages 847--864, New York, NY, USA, June 2019. Association for Computing Machinery.
\newblock ISBN 978-1-4503-5643-5.
\newblock \doi{10.1145/3299869.3300065}.
\newblock URL \url{https://dl.acm.org/doi/10.1145/3299869.3300065}.

\bibitem[Nargesian et~al.(2018)Nargesian, Zhu, Pu, and Miller]{nargesian_table_2018}
Fatemeh Nargesian, Erkang Zhu, Ken~Q. Pu, and Renée~J. Miller.
\newblock Table union search on open data.
\newblock \emph{Proceedings of the VLDB Endowment}, 11\penalty0 (7):\penalty0 813--825, March 2018.
\newblock ISSN 2150-8097.
\newblock \doi{10.14778/3192965.3192973}.
\newblock URL \url{https://dl.acm.org/doi/10.14778/3192965.3192973}.

\bibitem[Santos et~al.(2021)Santos, Bessa, Chirigati, Musco, and Freire]{santos_correlation_2021}
Aécio Santos, Aline Bessa, Fernando Chirigati, Christopher Musco, and Juliana Freire.
\newblock Correlation {Sketches} for {Approximate} {Join}-{Correlation} {Queries}.
\newblock In \emph{Proceedings of the 2021 {International} {Conference} on {Management} of {Data}}, pages 1531--1544, June 2021.
\newblock \doi{10.1145/3448016.3458456}.
\newblock URL \url{http://arxiv.org/abs/2104.03353}.
\newblock arXiv:2104.03353 [cs].

\bibitem[Srinivas et~al.(2023)Srinivas, Dolby, Abdelaziz, Hassanzadeh, Kokel, Khatiwada, Pedapati, Chaudhury, and Samulowitz]{srinivas_lakebench_2023}
Kavitha Srinivas, Julian Dolby, Ibrahim Abdelaziz, Oktie Hassanzadeh, Harsha Kokel, Aamod Khatiwada, Tejaswini Pedapati, Subhajit Chaudhury, and Horst Samulowitz.
\newblock {LakeBench}: {Benchmarks} for {Data} {Discovery} over {Data} {Lakes}, July 2023.
\newblock URL \url{http://arxiv.org/abs/2307.04217}.
\newblock arXiv:2307.04217 [cs].

\bibitem[Zhu et~al.(2017)Zhu, He, and Chaudhuri]{zhu_auto-join_2017}
Erkang Zhu, Yeye He, and Surajit Chaudhuri.
\newblock Auto-join: joining tables by leveraging transformations.
\newblock \emph{Proceedings of the VLDB Endowment}, 10\penalty0 (10):\penalty0 1034--1045, June 2017.
\newblock ISSN 2150-8097.
\newblock \doi{10.14778/3115404.3115409}.
\newblock URL \url{https://doi.org/10.14778/3115404.3115409}.

\bibitem[Cong et~al.(2023{\natexlab{a}})Cong, Gale, Frantz, Jagadish, and Demiralp]{cong_warpgate_2023}
Tianji Cong, James Gale, Jason Frantz, H.~V. Jagadish, and Çağatay Demiralp.
\newblock {WarpGate}: {A} {Semantic} {Join} {Discovery} {System} for {Cloud} {Data} {Warehouses}, January 2023{\natexlab{a}}.
\newblock URL \url{http://arxiv.org/abs/2212.14155}.
\newblock arXiv:2212.14155 [cs].

\bibitem[Cong et~al.(2023{\natexlab{b}})Cong, Nargesian, and Jagadish]{cong_pylon_2023}
Tianji Cong, Fatemeh Nargesian, and H.~V. Jagadish.
\newblock Pylon: {Semantic} {Table} {Union} {Search} in {Data} {Lakes}, January 2023{\natexlab{b}}.
\newblock URL \url{http://arxiv.org/abs/2301.04901}.
\newblock arXiv:2301.04901 [cs].

\bibitem[Zha et~al.(2023{\natexlab{b}})Zha, Zhou, Li, Wang, Huang, Yang, Yuan, Su, Li, Su, Zhang, Zhou, Shou, Wang, Zhu, Lu, Ye, Ye, Ye, Zhang, Deng, Xu, Wang, Chen, and Zhao]{zha_tablegpt_2023}
Liangyu Zha, Junlin Zhou, Liyao Li, Rui Wang, Qingyi Huang, Saisai Yang, Jing Yuan, Changbao Su, Xiang Li, Aofeng Su, Tao Zhang, Chen Zhou, Kaizhe Shou, Miao Wang, Wufang Zhu, Guoshan Lu, Chao Ye, Yali Ye, Wentao Ye, Yiming Zhang, Xinglong Deng, Jie Xu, Haobo Wang, Gang Chen, and Junbo Zhao.
\newblock {TableGPT}: {Towards} {Unifying} {Tables}, {Nature} {Language} and {Commands} into {One} {GPT}, August 2023{\natexlab{b}}.
\newblock URL \url{http://arxiv.org/abs/2307.08674}.
\newblock Issue: arXiv:2307.08674 arXiv:2307.08674 [cs].

\bibitem[Cheng et~al.(2023)Cheng, Li, and Bing]{cheng_is_2023}
Liying Cheng, Xingxuan Li, and Lidong Bing.
\newblock Is {GPT}-4 a {Good} {Data} {Analyst}?, May 2023.
\newblock URL \url{http://arxiv.org/abs/2305.15038}.
\newblock arXiv:2305.15038 [cs].

\bibitem[Zhang et~al.(2023)Zhang, Shen, Lu, and Zhuang]{zhang_data-copilot_2023}
Wenqi Zhang, Yongliang Shen, Weiming Lu, and Yueting Zhuang.
\newblock Data-{Copilot}: {Bridging} {Billions} of {Data} and {Humans} with {Autonomous} {Workflow}, June 2023.
\newblock URL \url{http://arxiv.org/abs/2306.07209}.
\newblock arXiv:2306.07209 [cs].

\bibitem[Li et~al.(2023)Li, Hui, Qu, Li, Yang, Li, Wang, Qin, Cao, Geng, Huo, Zhou, Ma, Li, Chang, Huang, Cheng, and Li]{li_can_2023}
Jinyang Li, Binyuan Hui, Ge~Qu, Binhua Li, Jiaxi Yang, Bowen Li, Bailin Wang, Bowen Qin, Rongyu Cao, Ruiying Geng, Nan Huo, Xuanhe Zhou, Chenhao Ma, Guoliang Li, Kevin C.~C. Chang, Fei Huang, Reynold Cheng, and Yongbin Li.
\newblock Can {LLM} {Already} {Serve} as {A} {Database} {Interface}? {A} {BIg} {Bench} for {Large}-{Scale} {Database} {Grounded} {Text}-to-{SQLs}, May 2023.
\newblock URL \url{http://arxiv.org/abs/2305.03111}.
\newblock arXiv:2305.03111 [cs].

\bibitem[Pourreza and Rafiei(2023)]{pourreza_din-sql_2023}
Mohammadreza Pourreza and Davood Rafiei.
\newblock {DIN}-{SQL}: {Decomposed} {In}-{Context} {Learning} of {Text}-to-{SQL} with {Self}-{Correction}, April 2023.
\newblock URL \url{http://arxiv.org/abs/2304.11015}.
\newblock arXiv:2304.11015 [cs].

\bibitem[Talmor et~al.(2021)Talmor, Yoran, Catav, Lahav, Wang, Asai, Ilharco, Hajishirzi, and Berant]{talmor_multimodalqa_2021}
Alon Talmor, Ori Yoran, Amnon Catav, Dan Lahav, Yizhong Wang, Akari Asai, Gabriel Ilharco, Hannaneh Hajishirzi, and Jonathan Berant.
\newblock {MultiModalQA}: {Complex} {Question} {Answering} over {Text}, {Tables} and {Images}, April 2021.
\newblock URL \url{http://arxiv.org/abs/2104.06039}.
\newblock arXiv:2104.06039 [cs].

\bibitem[Lin et~al.(2020)Lin, Socher, and Xiong]{lin_bridging_2020}
Xi~Victoria Lin, Richard Socher, and Caiming Xiong.
\newblock Bridging {Textual} and {Tabular} {Data} for {Cross}-{Domain} {Text}-to-{SQL} {Semantic} {Parsing}, December 2020.
\newblock URL \url{http://arxiv.org/abs/2012.12627}.
\newblock arXiv:2012.12627 [cs].

\bibitem[Zhang(2017)]{zhang_effective_2017}
Ziqi Zhang.
\newblock Effective and efficient {Semantic} {Table} {Interpretation} using {TableMiner}+.
\newblock \emph{Semantic Web}, 8\penalty0 (6):\penalty0 921--957, August 2017.
\newblock ISSN 22104968, 15700844.
\newblock \doi{10.3233/SW-160242}.
\newblock URL \url{https://www.medra.org/servlet/aliasResolver?alias=iospress&doi=10.3233/SW-160242}.

\bibitem[Parikh et~al.(2020)Parikh, Wang, Gehrmann, Faruqui, Dhingra, Yang, and Das]{parikh_totto_2020}
Ankur~P. Parikh, Xuezhi Wang, Sebastian Gehrmann, Manaal Faruqui, Bhuwan Dhingra, Diyi Yang, and Dipanjan Das.
\newblock {ToTTo}: {A} {Controlled} {Table}-{To}-{Text} {Generation} {Dataset}, October 2020.
\newblock URL \url{http://arxiv.org/abs/2004.14373}.
\newblock arXiv:2004.14373 [cs].

\bibitem[Korini and Bizer(2023)]{korini_column_2023}
Keti Korini and Christian Bizer.
\newblock Column {Type} {Annotation} using {ChatGPT}, July 2023.
\newblock URL \url{http://arxiv.org/abs/2306.00745}.
\newblock arXiv:2306.00745 [cs].

\bibitem[Yin et~al.(2020)Yin, Neubig, Yih, and Riedel]{yin_tabert_2020}
Pengcheng Yin, Graham Neubig, Wen-tau Yih, and Sebastian Riedel.
\newblock {TaBERT}: {Pretraining} for {Joint} {Understanding} of {Textual} and {Tabular} {Data}, May 2020.
\newblock URL \url{http://arxiv.org/abs/2005.08314}.
\newblock arXiv:2005.08314 [cs].

\bibitem[Deng et~al.(2020)Deng, Sun, Lees, Wu, and Yu]{deng_turl_2020}
Xiang Deng, Huan Sun, Alyssa Lees, You Wu, and Cong Yu.
\newblock {TURL}: table understanding through representation learning.
\newblock \emph{Proceedings of the VLDB Endowment}, 14\penalty0 (3):\penalty0 307--319, November 2020.
\newblock ISSN 2150-8097.
\newblock \doi{10.14778/3430915.3430921}.
\newblock URL \url{https://dl.acm.org/doi/10.14778/3430915.3430921}.

\bibitem[Tang et~al.(2021)Tang, Fan, Li, Tu, Du, Li, Madden, and Ouzzani]{tang_rpt_2021}
Nan Tang, Ju~Fan, Fangyi Li, Jianhong Tu, Xiaoyong Du, Guoliang Li, Sam Madden, and Mourad Ouzzani.
\newblock {RPT}: relational pre-trained transformer is almost all you need towards democratizing data preparation.
\newblock \emph{Proceedings of the VLDB Endowment}, 14\penalty0 (8):\penalty0 1254--1261, April 2021.
\newblock ISSN 2150-8097.
\newblock \doi{10.14778/3457390.3457391}.
\newblock URL \url{https://dl.acm.org/doi/10.14778/3457390.3457391}.

\bibitem[Herzig et~al.(2020)Herzig, Nowak, Müller, Piccinno, and Eisenschlos]{herzig_tapas_2020}
Jonathan Herzig, Paweł~Krzysztof Nowak, Thomas Müller, Francesco Piccinno, and Julian~Martin Eisenschlos.
\newblock {TAPAS}: {Weakly} {Supervised} {Table} {Parsing} via {Pre}-training.
\newblock In \emph{Proceedings of the 58th {Annual} {Meeting} of the {Association} for {Computational} {Linguistics}}, pages 4320--4333, 2020.
\newblock \doi{10.18653/v1/2020.acl-main.398}.
\newblock URL \url{http://arxiv.org/abs/2004.02349}.
\newblock arXiv:2004.02349 [cs].

\bibitem[Iida et~al.(2021)Iida, Thai, Manjunatha, and Iyyer]{iida_tabbie_2021}
Hiroshi Iida, Dung Thai, Varun Manjunatha, and Mohit Iyyer.
\newblock {TABBIE}: {Pretrained} {Representations} of {Tabular} {Data}, May 2021.
\newblock URL \url{http://arxiv.org/abs/2105.02584}.
\newblock arXiv:2105.02584 [cs].

\bibitem[McKinney(2010)]{mckinney_data_2010}
Wes McKinney.
\newblock Data {Structures} for {Statistical} {Computing} in {Python}.
\newblock In Stéfan van~der Walt and Jarrod Millman, editors, \emph{Proceedings of the 9th {Python} in {Science} {Conference}}, pages 56 -- 61, 2010.
\newblock \doi{10.25080/Majora-92bf1922-00a}.

\bibitem[Moritz et~al.(2018)Moritz, Nishihara, Wang, Tumanov, Liaw, Liang, Elibol, Yang, Paul, Jordan, and Stoica]{moritz_ray_2018}
Philipp Moritz, Robert Nishihara, Stephanie Wang, Alexey Tumanov, Richard Liaw, Eric Liang, Melih Elibol, Zongheng Yang, William Paul, Michael~I. Jordan, and Ion Stoica.
\newblock Ray: {A} {Distributed} {Framework} for {Emerging} {AI} {Applications}, September 2018.
\newblock URL \url{http://arxiv.org/abs/1712.05889}.
\newblock arXiv:1712.05889 [cs, stat].

\bibitem[Newman(2005)]{newman_power_2005}
M.~E.~J. Newman.
\newblock Power laws, {Pareto} distributions and {Zipf}'s law.
\newblock \emph{Contemporary Physics}, 46\penalty0 (5):\penalty0 323--351, September 2005.
\newblock ISSN 0010-7514, 1366-5812.
\newblock \doi{10.1080/00107510500052444}.
\newblock URL \url{http://arxiv.org/abs/cond-mat/0412004}.
\newblock Number: 5 arXiv:cond-mat/0412004.

\bibitem[Alstott et~al.(2014)Alstott, Bullmore, and Plenz]{alstott_powerlaw_2014}
Jeff Alstott, Ed~Bullmore, and Dietmar Plenz.
\newblock Powerlaw: a {Python} package for analysis of heavy-tailed distributions.
\newblock \emph{PLoS ONE}, 9\penalty0 (1):\penalty0 e85777, January 2014.
\newblock ISSN 1932-6203.
\newblock \doi{10.1371/journal.pone.0085777}.
\newblock URL \url{http://arxiv.org/abs/1305.0215}.
\newblock arXiv:1305.0215 [physics].

\bibitem[Johnson and Khoshgoftaar(2019)]{johnson_survey_2019}
Justin~M. Johnson and Taghi~M. Khoshgoftaar.
\newblock Survey on deep learning with class imbalance.
\newblock \emph{Journal of Big Data}, 6\penalty0 (1):\penalty0 27, March 2019.
\newblock ISSN 2196-1115.
\newblock \doi{10.1186/s40537-019-0192-5}.
\newblock URL \url{https://doi.org/10.1186/s40537-019-0192-5}.

\bibitem[Lee et~al.(2022)Lee, Ippolito, Nystrom, Zhang, Eck, Callison-Burch, and Carlini]{lee_deduplicating_2022}
Katherine Lee, Daphne Ippolito, Andrew Nystrom, Chiyuan Zhang, Douglas Eck, Chris Callison-Burch, and Nicholas Carlini.
\newblock Deduplicating {Training} {Data} {Makes} {Language} {Models} {Better}, March 2022.
\newblock URL \url{http://arxiv.org/abs/2107.06499}.
\newblock Issue: arXiv:2107.06499 arXiv:2107.06499 [cs].

\bibitem[Gao et~al.(2020)Gao, Biderman, Black, Golding, Hoppe, Foster, Phang, He, Thite, Nabeshima, Presser, and Leahy]{gao_pile_2020}
Leo Gao, Stella Biderman, Sid Black, Laurence Golding, Travis Hoppe, Charles Foster, Jason Phang, Horace He, Anish Thite, Noa Nabeshima, Shawn Presser, and Connor Leahy.
\newblock The {Pile}: {An} {800GB} {Dataset} of {Diverse} {Text} for {Language} {Modeling}, December 2020.
\newblock URL \url{http://arxiv.org/abs/2101.00027}.
\newblock Issue: arXiv:2101.00027 arXiv:2101.00027 [cs].

\bibitem[Mikolov et~al.(2013)Mikolov, Chen, Corrado, and Dean]{mikolov_efficient_2013}
Tomas Mikolov, Kai Chen, Greg Corrado, and Jeffrey Dean.
\newblock Efficient {Estimation} of {Word} {Representations} in {Vector} {Space}, September 2013.
\newblock URL \url{http://arxiv.org/abs/1301.3781}.
\newblock arXiv:1301.3781 [cs].

\bibitem[Reimers and Gurevych(2019)]{reimers_sentence-bert_2019}
Nils Reimers and Iryna Gurevych.
\newblock Sentence-{BERT}: {Sentence} {Embeddings} using {Siamese} {BERT}-{Networks}, August 2019.
\newblock URL \url{http://arxiv.org/abs/1908.10084}.
\newblock arXiv:1908.10084 [cs].

\end{thebibliography}

\clearpage

\section{Appendix}

\subsection{Sample Data}\label{sec:sample_data}
Below are some notable examples of tables in TabLib, including the table metadata and a random sample of the table, generated by loading the raw Arrow table, converting to Pandas, and formatting using \verb|df.sample(6).to_latex(index=False)|.


\subsubsection{Common Crawl: Salary Data from HTML Table}
Total Rows: 4

\begin{tabular}{rlll}
	\toprule
	Rank & Industry      & Average Salary & Hourly Rate \\
	\midrule
	3    & Professional  & \$32,844       & \$15.79     \\
	1    & Manufacturing & \$34,861       & \$16.76     \\
	4    & Retail        & \$32,337       & \$15.55     \\
	2    & Media         & \$34,771       & \$16.72     \\
	\bottomrule
\end{tabular}

\begin{minted}[frame=single,
				   framesep=3mm,
				   linenos=true,
				   xleftmargin=21pt,
				   tabsize=4,
				   fontsize=\tiny,
				   breaklines=true,
				   breakanywhere=true]{js}
{
  "warc_record_id": "<urn:uuid:f037aaa0-c9dd-4e63-be55-28206b8aae4a>",
  "warc_target_uri": "https://www.zippia.com/merchandising-specialist-jobs/salary/",
  "warc_date": "2023-06-07T14:51:22Z",
  "warc_path": "crawl-data/CC-MAIN-2023-23/segments/1685224653930.47/warc/CC-MAIN-20230607143116-20230607173116-00333.warc.gz",
  "extractor": "html",
  "html_title": "Merchandising Specialist Salary (June 2023) - Zippia",
  "html_metadata": {
    "og": {
      "title": "Merchandising Specialist Salary (June 2023) - Zippia",
      "type": "website",
      "url": "https://www.zippia.com/merchandising-specialist-jobs/salary/",
      "description": "The average salary for a Merchandising Specialist is $32,000 per year, or $16 per hour in United States. Find out the average a salary by state, years of experience, and field.",
      "updated_time": "2023-04-06T02:00:00-08:00",
      "image": "https://static.zippia.com/assets/zippia-og-image.png"
    },
    "meta": {
      "viewport": "height=device-height, width=device-width, initial-scale=1.0, viewport-fit=cover",
      "description": "The average salary for a Merchandising Specialist is $32,000 per year, or $16 per hour in United States. Find out the average a salary by state, years of experience, and field.",
      "author": "",
      "og:title": "Merchandising Specialist Salary (June 2023) - Zippia",
      "og:type": "website",
      "og:url": "https://www.zippia.com/merchandising-specialist-jobs/salary/",
      "og:description": "The average salary for a Merchandising Specialist is $32,000 per year, or $16 per hour in United States. Find out the average a salary by state, years of experience, and field.",
      "article:published_time": "2020-05-18T00:00:00-08:00",
      "article:modified_time": "2023-04-06T02:00:00-08:00",
      "og:updated_time": "2023-04-06T02:00:00-08:00",
      "twitter:card": "summary",
      "twitter:site": "@ZippiaInc",
      "twitter:title": "Merchandising Specialist Salary (June 2023) - Zippia",
      "twitter:url": "https://www.zippia.com/merchandising-specialist-jobs/salary/",
      "twitter:description": "The average salary for a Merchandising Specialist is $32,000 per year, or $16 per hour in United States. Find out the average a salary by state, years of experience, and field.",
      "charset": [
        "utf8",
        "utf-8"
      ],
      "twitter:image:src": "https://static.zippia.com/assets/zippia-og-image.png",
      "fb:app_id": "508633732650088",
      "og:image": "https://static.zippia.com/assets/zippia-og-image.png",
      "next-head-count": "24",
      "X-UA-Compatible": "IE=edge"
    },
    "dc": {},
    "page": {
      "title": "Merchandising Specialist Salary (June 2023) - Zippia",
      "canonical": "https://www.zippia.com/merchandising-specialist-jobs/salary/"
    },
    "twitter": {
      "card": "summary",
      "site": "@ZippiaInc",
      "title": "Merchandising Specialist Salary (June 2023) - Zippia",
      "url": "https://www.zippia.com/merchandising-specialist-jobs/salary/",
      "description": "The average salary for a Merchandising Specialist is $32,000 per year, or $16 per hour in United States. Find out the average a salary by state, years of experience, and field.",
      "image:src": "https://static.zippia.com/assets/zippia-og-image.png"
    },
    "_internal": {
      "url": null,
      "url_actual": null
    },
    "_v": 1
  },
  "before": "18.33\n5\n7\nVera Bradley\n$38,023\n$18.28\n8\nClover Food Lab\n$36,984\n$17.78\n9\nThe Mosaic Company\n$36,833\n$17.71\n5\n10\nAnheuser-Busch\n$36,595\n$17.59\n22\n11\nGreen Mountain Coffee Roasters\n$36,516\n$17.56\n12\nSWFcontract\n$36,120\n$17.37\n13\nTarget\n$35,993\n$17.30\n75\n14\nGap Inc.\n$35,806\n$17.21\n4\n15\nNintendo\n$35,651\n$17.14\n16\nMichael Kors\n$35,636\n$17.13\n2\n17\nYaamava' Resort & Casino\n$35,515\n$17.07\n1\n18\nEmpire Cat\n$35,189\n$16.92\n19\nAmerisourceBergen\n$35,170\n$16.91\n20\nSummer Classics\n$35,012\n$16.83\nShow More\nHow Much Do Merchandising Specialists Make In Different Industries?\nHere are some examples of how much a merchandising specialist salaries can based on different industries:\nThe manufacturing industry pays merchandising specialists an average salary of $34,861\nThe media industry pay $34,771\nThe lowest paying industry for merchandising specialists is the retail industry. Merchandising specialists in this industry earn an average salary of $32,337\nHighest Paying Industries For Merchandising Specialists",
  "after": "High Paying Merchandising Specialist Jobs\nMerchandising Specialist Pay Trends\nAverage Merchandising Specialist Salary Over Time\nCompare salaries for individual cities or states with the national average.\nAshburn, VA\nMerchandising Specialist Salary By Year\nYear\nAvg. Salary\nHourly Rate\n% Change\n2023\n$32,900\n$15.83\n+4.4%\n2022\n$31,500\n$15.13\n+3.4%\n2021\n$30,400\n$14.61\n+2.7%\n2020\n$29,600\n$14.22\n+3.5%\n2019\n$28,500\n$13.72\n-0.6%\nShow More\nRecently Added Merchandising Specialist Salaries\nAshburn, VA\nCompany\nJob\nLocation\nDate Added\nSalary\nBDS Connected Solutions\nRetail Merchandising Specialist\nDenver, CO\n05/04/2023\n$33,392\nFloor & Decor\nMerchandise Specialist\nDanbury, CT\n05/02/2023\n$33,392\nRecruiter Relentless\nTerritory Merchandising Specialist-Grocery\nSanta Barbara, CA\n05/02/2023\n$37,566\nZinus Inc.\nSite Merchandising Specialist\nTracy, CA\n05/01/2023\n$70,000\nAcosta, Inc.\nRetail Merchandising Specialist\nNew York, NY\n04/30/2023\n$33,392\nAcosta, Inc.\nRetail Merchandising Specialist\nSeattle, WA\n04/30/2",
  "mime_type": "text/html"
}

\end{minted}

\subsubsection{GitHub: COVID Data from Excel Spreadsheet}
Total Rows: 1099

\begin{tabular}{lrrr}
	\toprule
	Date       & Positive Total & Positive New & 7-day confirmed case average \\
	\midrule
	2022-10-13 & 1900373        & 1246         & 1083.000000                  \\
	2020-12-06 & 257244         & 2094         & 4678.142857                  \\
	2023-01-02 & 1979199        & 1420         & 1426.857143                  \\
	2020-06-08 & 99460          & 349          & 324.571429                   \\
	2021-04-29 & 645210         & 998          & 1073.428571                  \\
	2020-09-23 & 127148         & 558          & 375.571429                   \\
	\bottomrule
\end{tabular}
\begin{minted}[frame=single,
		framesep=3mm,
		linenos=true,
		xleftmargin=21pt,
		tabsize=4,
		fontsize=\tiny,
		breaklines=true,
		breakanywhere=true]{js}

{
	"github_repo": "BioTurboNick/MassCovid.jl",
	"github_ref": "refs/heads/master",
	"github_hash": "65cf916",
	"github_repo_path": "input/february-2-2023.xlsx",
	"excel_sheet": "CasesByDate (Test Date)",
	"excel_other_sheets": [
		"Data Documentation",
		"Weekly_Town_Reference",
		"Age Means Last2Weeks",
		"AgeLast2Weeks",
		"Cases (Report Date)",
		"CasesbyAge",
		"CasesByDate_Probable",
		"CountyCasesDeaths (Report Date)",
		"County_Weekly",
		"CountyDeaths",
		"DateofDeath",
		"DeathsReported (Report Date)",
		"DeathCharacteristics",
		"HigherEd_CasesandTests",
		"HospBed-Hospital COVID Census",
		"HospBedAvailable-Regional",
		"New Hospital Demographic Data",
		"Hospitalization from Hospitals",
		"LTC Facilities",
		"RaceEthnicityLast2Weeks",
		"SexLast2Weeks",
		"Testing2 (Report Date)",
		"TestingByDate (Test Date)",
		"TestingPosByAge",
		"Weekly_City_Town",
		"Weekly_Statewide",
		"Clusters",
		"Isolation and Quarantine",
		"Contact Tracing",
		"CTC workforce",
		"Counts by Specimen Date (Sero)"
	],
	"extractor": "excel",
	"mime_type": "application/vnd.openxmlformats-officedocument.spreadsheetml.sheet",
	"tar_path": "BioTurboNick-MassCovid.jl-65cf916/input/february-2-2023.xlsx"
}
\end{minted}

\subsubsection{GitHub: CDC Report from 2000 (BRFSS) Extracted From PDF }
Total Rows: 53

\begin{tabular}{llll}
	\toprule
	State        & BRFSS Percent & Population Percent & Difference \\
	\midrule
	Louisiana    & 18.51         & 19.28              & -0.77      \\
	Oregon       & 21.13         & 17.30              & 3.83       \\
	Maine        & 16.95         & 17.16              & -0.21      \\
	New York     & 20.88         & 18.41              & 2.47       \\
	Nevada       & 21.94         & 18.65              & 3.29       \\
	North Dakota & 15.85         & 17.69              & -1.84      \\
	\bottomrule
\end{tabular}
\begin{minted}[frame=single,
	framesep=3mm,
	linenos=true,
	xleftmargin=21pt,
	tabsize=4,
	fontsize=\tiny,
	breaklines=true,
	breakanywhere=true]{js}
{
	"github_repo": "dkastner/cdc-backup-data",
	"github_ref": "refs/heads/master",
	"github_hash": "bb00f0c",
	"github_repo_path": "www.cdc.gov/brfss/annual_data/2000/pdf/2000summarydataqualityreport.pdf",
	"extractor": "pdf",
	"pdf_bbox": [
		34.50001000000001,
		62.279964999999976,
		502.49996500000134,
		696.71999125
	],
	"pdf_page": 15,
	"pdf_metadata": {
		"Author": "CDC",
		"Company": "CDC",
		"CreationDate": "D:20040406165511-04'00'",
		"Creator": "Acrobat PDFMaker 6.0 for Word",
		"ModDate": "D:20140127134602-05'00'",
		"Producer": "Acrobat Distiller 6.0 (Windows)",
		"SourceModified": "D:20040406204057",
		"Title": "2000 BRFSS Summary Data Quality Control Report"
	},
	"before": "higan 11.98 13.11 -1.14\nNew Mexico 11.60 12.76 -1.16\nOklahoma 11.79 12.95 -1.16\nDelaware 11.91 13.24 -1.33\nIllinois 11.29 12.63 -1.33\nMontana 10.60 11.98 -1.38\nMissouri 11.04 12.46 -1.43\nNew Jersey 9.44 10.95 -1.51\nConnecticut 9.62 11.32 -1.71\nMinnesota 10.50 12.25 -1.74\nArkansas 10.70 12.69 -1.99\nPennsylvania 10.08 12.15 -2.07\nSouth Dakota 10.96 13.16 -2.21\nOhio 10.49 12.78 -2.29\nSouth Carolina 11.61 13.90 -2.29\nNorth Carolina 10.99 13.33 -2.33\nTennessee 10.38 12.73 -2.36\nRhode Island 10.59 13.13 -2.55\nWest Virginia 9.94 12.51 -2.58\nAlabama 10.76 13.43 -2.67\nGeorgia 10.62 13.37 -2.75\nMassachusetts 10.00 12.75 -2.75\nKentucky 10.25 13.12 -2.87\nMississippi 11.71 14.76 -3.05\nIowa 10.06 13.15 -3.09\nVirginia 9.59 12.88 -3.29\nVermont 10.41 13.90 -3.48\nIndiana 9.78 13.46 -3.68\nMaine 7.57 12.44 -4.87\nPuerto Rico 11.73 17.28 -5.56\nWisconsin 6.91 12.82 -5.91\nNew Hampshire 6.79 12.75 -5.95\nMedian 11.08 12.77 -1.41Table 9. Percentage of People Aged 25\u201334 in BRFSS and Population Data by State, 2000",
	"after": "Table 10. Percentage of People Aged 35\u201344 in BRFSS and Population Data by State, 2000\nState BRFSS Percent Population Percent Difference\nNew Hampshire 27.28 22.13 5.14\nVirginia 26.21 22.00 4.21\nNew Jersey 25.09 21.35 3.74\nWisconsin 24.23 20.87 3.37\nNew York 24.26 21.14 3.12\nIowa 22.14 19.29 2.85\nSouth Dakota 22.64 19.96 2.68\nPennsylvania 22.38 19.88 2.50\nRhode Island 23.44 20.97 2.47\nConnecticut 23.99 21.59 2.41\nMississippi 22.03 19.75 2.28\nOhio 22.69 20.45 2.24\nIllinois 23.40 21.22 2.18\nMassachusetts 23.28 21.16 2.12\nMinnesota 23.33 21.24 2.09\nIndiana 22.51 20.46 2.04\nAlabama 21.69 19.87 1.82\nNorth Dakota 21.77 20.04 1.72\nIdaho 22.22 20.64 1.58\nFlorida 20.85 19.35 1.50\nHawaii 22.38 20.92 1.46\nMissouri 21.71 20.35 1.36\nDelaware 22.00 20.89 1.11\nSouth Carolina 21.71 20.65 1.06\nWest Virginia 19.78 18.73 1.05\nWyoming 22.98 21.94 1.04\nTennessee 21.51 20.49 1.03\nGeorgia 23.05 22.05 1.00\nMaine 22.01 21.07 0.94\nVermont 22.50 21.56 0.94\nMontana 21.52 20.68 0.84\nNew Mexico 22.81 22.00 0.80\nKansa",
	"mime_type": "application/pdf",
	"tar_path": "erithmetic-cdc-backup-data-bb00f0c/www.cdc.gov/brfss/annual_data/2000/pdf/2000summarydataqualityreport.pdf"
}
\end{minted}
\subsubsection{Common Crawl: HTML Calendar}
These HTML-formatted calendars occur frequently in the Common Crawl dataset.

Total Rows: 6

\CJK{UTF8}{gbsn}
\begin{tabular}{lllrrrr}
	\toprule
	日    & 月    & 火    & 水    & 木    & 金    & 土    \\
	\midrule
	11   & 12   & 13   & 14.0 & 15.0 & 16.0 & 17.0 \\
	None & None & None & NaN  & 1.0  & 2.0  & 3.0  \\
	25   & 26   & 27   & 28.0 & 29.0 & 30.0 & NaN  \\
	« 4月 & « 4月 & « 4月 & NaN  & NaN  & NaN  & NaN  \\
	18   & 19   & 20   & 21.0 & 22.0 & 23.0 & 24.0 \\
	4    & 5    & 6    & 7.0  & 8.0  & 9.0  & 10.0 \\
	\bottomrule
\end{tabular}
\begin{minted}[frame=single,
		framesep=3mm,
		linenos=true,
		xleftmargin=21pt,
		tabsize=4,
		fontsize=\tiny,
		breaklines=true,
		breakanywhere=true]{js}
{
	"warc_record_id": "<urn:uuid:8073c7fb-b523-4792-b1c0-936f7e1742a0>",
	"warc_target_uri": "https://www.arimatsu-dental.jp/news/354/attachment/%E6%AD%AF%E5%8C%BB%E8%80%85%E6%BC%AB%E7%94%BB1-1-3",
	"warc_date": "2023-06-04T15:03:06Z",
	"warc_path": "crawl-data/CC-MAIN-2023-23/segments/1685224649986.95/warc/CC-MAIN-20230604125132-20230604155132-00518.warc.gz",
	"extractor": "html",
	"html_title": "\u00bb \u6b6f\u533b\u8005\u6f2b\u753b1-1",
	"html_metadata": {
	"og": {},
	"meta": {
		"charset": "utf-8",
		"X-UA-Compatible": "IE=edge,chrome=1",
		"viewport": "width=device-width, initial-scale=1"
	},
	"dc": {},
	"page": {
		"title": "\u00bb \u6b6f\u533b\u8005\u6f2b\u753b1-1",
		"shortlink": "https://www.arimatsu-dental.jp/?p=367"
	},
	"twitter": {},
	"_internal": {
		"url": null,
		"url_actual": null
	},
	"_v": 1
	},
	"before": ";\n    js = d.createElement(s); js.id = id;\n    js.src = \"//connect.facebook.net/ja_JP/sdk.js#xfbml=1&version=v2.0\";\n    fjs.parentNode.insertBefore(js, fjs);\n  }(document, 'script', 'facebook-jssdk'));\n\u521d\u3081\u3066\u306e\u65b9\nbr><span style=\"font-size: 12px;\">LINE\u3067\u3054\u4e88\u7d04</span\n\u901a\u9662\u4e2d\u306e\u65b9\nLINE\u3067\u3054\u4e88\u7d04\nHOME\n\u8a3a\u7642\u6848\u5185\n\u8a3a\u7642\u6848\u5185\n\u6b6f\u5468\u75c5\u306b\u3064\u3044\u3066\n\u30af\u30e9\u30a6\u30f3\uff08\u88ab\u305b\u7269\uff09\u306e\u30e1\u30cb\u30e5\u30fc\n\u90e8\u5206\u5165\u308c\u6b6f\u306e\u30e1\u30cb\u30e5\u30fc\n\u7dcf\u5165\u308c\u6b6f\u306e\u30e1\u30cb\u30e5\u30fc\n\u6b6f\u79d1\u76f8\u8ac7\n\u533b\u9662\u6848\u5185\nsearch\n\u30e1\u30cb\u30e5\u30fc\u958b\u9589\n\u6b6f\u533b\u8005\u6f2b\u753b1-1\n\u521d\u3081\u3066\u306e\u65b9\n\u901a\u9662\u4e2d\u306e\u65b9\nTOP\n\u6b6f\u533b\u8005\u6f2b\u753b1-1\n2018-05-07\nTweet\n\u26054\u30b3\u30de\u6f2b\u753b\u66f4\u65b0\u2605\nYou can start editing here.\nIf comments are open, but there are no comments.\n\u30b3\u30e1\u30f3\u30c8\u3092\u6b8b\u3059\n\u30b3\u30e1\u30f3\u30c8\u3092\u30ad\u30e3\u30f3\u30bb\u30eb\n\u30e1\u30fc\u30eb\u30a2\u30c9\u30ec\u30b9\u304c\u516c\u958b\u3055\u308c\u308b\u3053\u3068\u306f\u3042\u308a\u307e\u305b\u3093\u3002\n*\n\u304c\u4ed8\u3044\u3066\u3044\u308b\u6b04\u306f\u5fc5\u9808\u9805\u76ee\u3067\u3059\n\u30b3\u30e1\u30f3\u30c8\n\u540d\u524d\n*\n\u30e1\u30fc\u30eb\u30a2\u30c9\u30ec\u30b9\n*\n\u30b5\u30a4\u30c8\n#respond\n\u30da\u30fc\u30b8\u4e00\u89a7\nHOME\n\u8a3a\u7642\u6848\u5185\n\u6b6f\u5468\u75c5\u306b\u3064\u3044\u3066\n\u30af\u30e9\u30a6\u30f3\uff08\u88ab\u305b\u7269\uff09\u306e\u30e1\u30cb\u30e5\u30fc\n\u90e8\u5206\u5165\u308c\u6b6f\u306e\u30e1\u30cb\u30e5\u30fc\n\u7dcf\u5165\u308c\u6b6f\u306e\u30e1\u30cb\u30e5\u30fc\n\u6b6f\u79d1\u76f8\u8ac7\uff31\uff06\uff21\n\u533b\u9662\u6848\u5185\n\u30d7\u30e9\u30a4\u30d0\u30b7\u30fc\u30dd\u30ea\u30b7\u30fc\n\u30b5\u30a4\u30c8\u30de\u30c3\u30d7\n\u30ab\u30c6\u30b4\u30ea\u30fc\n4\u30b3\u30de\u6f2b\u753b\n\u304a\u77e5\u3089\u305b\n\u6700\u8fd1\u306e\u6295\u7a3f\n\uff16\u6708\u306e\u4f11\u8a3a\u65e5\u306b\u3064\u3044\u3066\nGW\u306e\u4f11\u8a3a\u65e5\u306b\u3064\u3044\u3066\n4\u30fb5\u6708\u306e\u4f11\u8a3a\u65e5\u306b\u3064\u3044\u3066\n\u6765\u9662\u3055\u308c\u308b\u7686\u69d8\u3078\n\u30aa\u30f3\u30e9\u30a4\u30f3\u8cc7\u683c\u78ba\u8a8d\u306b\u3064\u3044\u3066\n\u30a2\u30fc\u30ab\u30a4\u30d6\n2023\u5e744\u6708\n2023\u5e743\u6708\n2023\u5e742\u6708\n2023\u5e741\u6708\n2022\u5e7411\u6708\n2022\u5e749\u6708\n2022\u5e748\u6708\n2022\u5e745\u6708\n2022\u5e744\u6708\n2022\u5e742\u6708\n2022\u5e741\u6708\n2021\u5e7410\u6708\n2021\u5e746\u6708\n2021\u5e744\u6708\n2020\u5e7410\u6708\n2020\u5e744\u6708\n2019\u5e743\u6708\n2019\u5e742\u6708\n2019\u5e741\u6708\n2018\u5e7412\u6708\n2018\u5e7411\u6708\n2018\u5e7410\u6708\n2018\u5e749\u6708\n2018\u5e748\u6708\n2018\u5e747\u6708\n2018\u5e746\u6708\n2018\u5e745\u6708\n2018\u5e744\u6708\n2018\u5e743\u6708",
	"after": "1\n\u30d7\u30e9\u30a4\u30d0\u30b7\u30fc\u30dd\u30ea\u30b7\u30fc\n\u30b5\u30a4\u30c8\u30de\u30c3\u30d7\nCopyright \u00a9 \u3042\u308a\u307e\u3064\u6b6f\u79d1 All Rights Reserved.\n\u3010\u63b2\u8f09\u306e\u8a18\u4e8b\u30fb\u5199\u771f\u30fb\u30a4\u30e9\u30b9\u30c8\u306a\u3069\u306e\u7121\u65ad\u8907\u5199\u30fb\u8ee2\u8f09\u7b49\u3092\u7981\u3058\u307e\u3059\u3011\ntwitter\n!function(d,s,id){var js,fjs=d.getElementsByTagName(s)[0],p=/^http:/.test(d.location)?'http':'https';if(!d.getElementById(id)){js=d.createElement(s);js.id=id;js.src=p+'://platform.twitter.com/widgets.js';fjs.parentNode.insertBefore(js,fjs);}}(document, 'script', 'twitter-wjs');\ngoogle+\n{lang: \"ja\"}\n/* <![CDATA[ */\nvar wpcf7 = {\"apiSettings\":{\"root\":\"https:\\/\\/www.arimatsu-dental.jp\\/wp-json\\/contact-form-7\\/v1\",\"namespace\":\"contact-form-7\\/v1\"},\"recaptcha\":{\"messages\":{\"empty\":\"\\u3042\\u306a\\u305f\\u304c\\u30ed\\u30dc\\u30c3\\u30c8\\u3067\\u306f\\u306a\\u3044\\u3053\\u3068\\u3092\\u8a3c\\u660e\\u3057\\u3066\\u304f\\u3060\\u3055\\u3044\\u3002\"}}};\n/* ]]> */",
	"mime_type": "text/html"
}
\end{minted}
\subsubsection{Duplicated Data Example}
\label{sec:duplicate-example}
Below is an example of duplicated tables (based on content hash), that occur within different GitHub repositories, but have the same test file (a standard introductory machine learning Titantic dataset). Other common occurrences of duplication include: same repository but different paths (i.e. many sub-folders with same files), different repositories but same code/files/documentation, and so on.

We've shown the first 7 rows and first 6 columns of the table below, which has a couple interesting characteristics:

\begin{itemize}
    \item They were parsed as HTML from a \verb|.ipynb| file, in the output of a Jupyter notebook. Because the \verb|.ipynb| extension is unknown to our parser, we inspected the file with \verb|libmagic| which classified the file as HTML.
    \item The 6th row literally contains ``...'', because \verb|df.__str__()| truncates the middle section of a dataframe when printing it. This is part of the parsed table and was not added for display.
\end{itemize}

Total Rows: 11

\begin{tabular}{llllll}
\toprule
PassengerId & Survived & Pclass &                                              Name &    Sex &  Age \\
\midrule
          1 &        0 &      3 &                           Braund, Mr. Owen Harris &   male & 22.0 \\
          2 &        1 &      1 & Cumings, Mrs. John Bradley (Florence Briggs Th... & female & 38.0 \\
          3 &        1 &      3 &                            Heikkinen, Miss. Laina & female & 26.0 \\
          4 &        1 &      1 &      Futrelle, Mrs. Jacques Heath (Lily May Peel) & female & 35.0 \\
          5 &        0 &      3 &                          Allen, Mr. William Henry &   male & 35.0 \\
        ... &      ... &    ... &                                               ... &    ... &  ... \\
        887 &	0 &	2 &	Montvila, Rev. Juozas &	male &	27.0 \\
\bottomrule
\end{tabular}

Context metadata for first source:

\begin{minted}[frame=single,
		framesep=3mm,
		linenos=true,
		xleftmargin=21pt,
		tabsize=4,
		fontsize=\tiny,
		breaklines=true,
		breakanywhere=true]{js}
{
  "github_repo": "mdmiqbal/Titanic-dataset",
  "github_ref": "refs/heads/main",
  "github_hash": "d80f02d",
  "github_repo_path": "Assign 1(titenic data set 0).ipynb",
  "extractor": "html",
  "sourceline": 129,
  "sourcepos": 8,
  "before": "\\n\",\n       \"\\n\",\n       \"    .dataframe tbody tr th:only-of-type {\\n\",\n       \"        vertical-align: middle;\\n\",\n       \"    }\\n\",\n       \"\\n\",\n       \"    .dataframe tbody tr th {\\n\",\n       \"        vertical-align: top;\\n\",\n       \"    }\\n\",\n       \"\\n\",\n       \"    .dataframe thead th {\\n\",\n       \"        text-align: right;\\n\",\n       \"    }\\n\",\n       \"\\n\",\n       \"",
  "after": "\\n\",\n       \"891 rows \u00d7 12 columns\\n\",\n       \"",
  "mime_type": "text/html",
  "tar_path": "mdmiqbal-Titanic-dataset-d80f02d/Assign 1(titenic data set 0).ipynb"
}
\end{minted}

Context metadata for second source:

\begin{minted}[frame=single,
		framesep=3mm,
		linenos=true,
		xleftmargin=21pt,
		tabsize=4,
		fontsize=\tiny,
		breaklines=true,
		breakanywhere=true]{js}
{
  "github_repo": "nkaraffa/Intro-to-AI-Machine-Learning-and-Python-basics",
  "github_ref": "refs/heads/main",
  "github_hash": "c45317a",
  "github_repo_path": "Classification_Model_Titanic.ipynb",
  "extractor": "html",
  "sourceline": 131,
  "sourcepos": 15,
  "before": "-type {\\n\",\n              \"        vertical-align: middle;\\n\",\n              \"    }\\n\",\n              \"\\n\",\n              \"    .dataframe tbody tr th {\\n\",\n              \"        vertical-align: top;\\n\",\n              \"    }\\n\",\n              \"\\n\",\n              \"    .dataframe thead th {\\n\",\n              \"        text-align: right;\\n\",\n              \"    }\\n\",\n              \"\\n\",\n              \"",
  "after": "\\n\",\n              \"891 rows \u00c3\u2014 12 columns\\n\",\n              \"",
  "mime_type": "text/html",
  "tar_path": "nkaraffa-Intro-to-AI-Machine-Learning-and-Python-basics-c45317a/Classification_Model_Titanic.ipynb"
}
\end{minted}

\subsubsection{Unknown Language Example}
\label{sec:unknown-lang-example}
Below is an example of a table which was classified as "Unknown" language. This particular table was entirely numeric, providing no language hints.

\begin{tabular}{rrrrrrrrrrr}
\toprule
 2.0 &      0.0 &  0.6363636363636364 &      1.0 &  0.0.1 &    0.0.2 &  0.0.3 &  1.0.1 &  0.0.4 &  1.0.2 &  0.0.5 \\
\midrule
 2.0 & 0.000000 &            0.909091 & 1.000000 &    0.0 & 0.000000 &    0.0 &    1.0 &    0.0 &    1.0 &    0.0 \\
 2.0 & 0.000000 &            1.000000 & 1.000000 &    0.0 & 0.000000 &    0.0 &    1.0 &    0.0 &    1.0 &    0.0 \\
 3.0 & 0.842857 &            0.085714 & 0.101695 &    0.0 & 0.030928 &    0.0 &    0.0 &    1.0 &    0.0 &    1.0 \\
 3.0 & 0.891892 &            0.027027 & 0.030303 &    0.0 & 0.020619 &    0.0 &    0.0 &    1.0 &    0.0 &    1.0 \\
 2.0 & 0.000000 &            0.909091 & 1.000000 &    0.0 & 0.000000 &    0.0 &    1.0 &    0.0 &    1.0 &    0.0 \\
 3.0 & 0.850000 &            0.075000 & 0.088235 &    0.0 & 0.010309 &    0.0 &    1.0 &    0.0 &    0.0 &    1.0 \\
\bottomrule
\end{tabular}

\begin{minted}[frame=single,
		framesep=3mm,
		linenos=true,
		xleftmargin=21pt,
		tabsize=4,
		fontsize=\tiny,
		breaklines=true,
		breakanywhere=true]{js}
{
  "github_repo": "nkaraffa/Intro-to-AI-Machine-Learning-and-Python-basics",
  "github_ref": "refs/heads/main",
  "github_hash": "c45317a",
  "github_repo_path": "Classification_Model_Titanic.ipynb",
  "extractor": "html",
  "sourceline": 131,
  "sourcepos": 15,
  "before": "-type {\\n\",\n              \"        vertical-align: middle;\\n\",\n              \"    }\\n\",\n              \"\\n\",\n              \"    .dataframe tbody tr th {\\n\",\n              \"        vertical-align: top;\\n\",\n              \"    }\\n\",\n              \"\\n\",\n              \"    .dataframe thead th {\\n\",\n              \"        text-align: right;\\n\",\n              \"    }\\n\",\n              \"\\n\",\n              \"",
  "after": "\\n\",\n              \"891 rows \u00c3\u2014 12 columns\\n\",\n              \"",
  "mime_type": "text/html",
  "tar_path": "nkaraffa-Intro-to-AI-Machine-Learning-and-Python-basics-c45317a/Classification_Model_Titanic.ipynb"
}
\end{minted}

\end{document}